\documentclass{article}

\usepackage{microtype}
\usepackage{graphicx}
\usepackage{subcaption}
\usepackage{booktabs} %

\usepackage{hyperref}

\usepackage[table]{xcolor}
\usepackage{amssymb}

\usepackage[preprint]{icml2026}

\usepackage{amsmath}
\usepackage{amssymb}
\usepackage{mathtools}
\usepackage{amsthm}
\usepackage{CJKutf8}

\usepackage{multirow}

\usepackage{xcolor}         %

\usepackage[capitalize,noabbrev]{cleveref}

\theoremstyle{plain}

\theoremstyle{definition}

\theoremstyle{remark}

\usepackage[textsize=tiny]{todonotes}

\icmltitlerunning{Efficient Decoder Scaling Strategy for Neural Routing Solvers}

\begin{document}
\begin{CJK*}{UTF8}{gbsn}

\twocolumn[
  \icmltitle{Efficient Decoder Scaling Strategy for Neural Routing Solvers}

  \icmlsetsymbol{equal}{*}

  \begin{icmlauthorlist}
    \icmlauthor{Qing Luo}{yyy}
    \icmlauthor{Fu Luo}{yyy}
    \icmlauthor{Ke Li}{yyy,comp}
    \icmlauthor{Zhenkun Wang}{yyy}
  \end{icmlauthorlist}

  \icmlaffiliation{yyy}{Guangdong Provincial Key Laboratory of Fully Actuated System Control Theory and Technology, School of Automation and Intelligent Manufacturing, Southern University of Science and Technology, Shenzhen, China.}
  \icmlaffiliation{comp}{Department of Computer Science, City University of Hong Kong, Hong Kong SAR, China}

  \icmlcorrespondingauthor{Zhenkun Wang}{wangzhenkun90@gmail.com}

  \icmlkeywords{Machine Learning, ICML}

  \vskip 0.3in
]

\printAffiliationsAndNotice{}  %

\begin{abstract}

Construction-based neural routing solvers, typically composed of an encoder and a decoder, have emerged as a promising approach for solving vehicle routing problems. While recent studies suggest that shifting parameters from the encoder to the decoder enhances performance, most works restrict the decoder size to 1–3M parameters, leaving the effects of scaling largely unexplored. To address this gap, we conduct a systematic study comparing two distinct strategies: scaling depth versus scaling width. We synthesize these strategies to construct a suite of 12 model configurations, spanning a parameter range from 1M to $\sim$150M, and extensively evaluate their scaling behaviors across three critical dimensions: parameter efficiency, data efficiency, and compute efficiency. Our empirical results reveal that parameter count is insufficient to accurately predict the model performance, highlighting the critical and distinct roles of model depth (layer count) and width (embedding dimension). Crucially, we demonstrate that scaling depth yields superior performance gains to scaling width. Based on these findings, we provide and experimentally validate a set of design principles for the efficient allocation of parameters and compute resources to enhance the model performance.

\end{abstract}

\begin{figure}[t]
    \centering
    \includegraphics[width=1.0\linewidth]{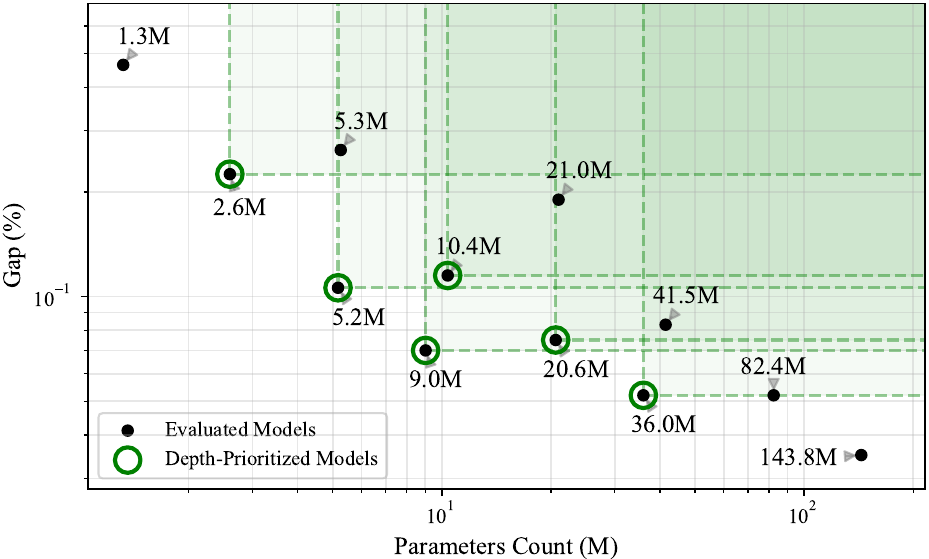}
    \caption{Decoder Parameters Count vs. Optimality Gap. The results demonstrate that simply increasing model size does not guarantee performance gains. The green circles highlight depth-prioritized models (constructed by prioritizing depth scaling over width), which dominate at least one larger model.}
    \label{fig:intro_efficient models}
\end{figure}

\section{Introduction}

The Traveling Salesman Problem (TSP) is a classic NP-hard problem in Combinatorial Optimization (CO) and is widely applied in various domains such as logistics~\citep{konstantakopoulos2022vehicle} and chip manufacturing~\citep{brophy2014principles}. Real-world scenarios often rely on specialized heuristics, which require substantial domain knowledge. Recent developments in Neural Combinatorial Optimization (NCO) leverage deep learning to enable end-to-end neural solvers to solve CO problems~\citep{vinyals2015pointer,bengio2021machine}. A key advantage of these approaches is their ability to learn heuristics automatically from data, thereby reducing the burden of expert-driven design. 

Construction-based methods represent one of the mainstream paradigms in NCO, which builds solutions autoregressively by sequentially selecting unvisited nodes until a complete solution is formed. The node selection process is guided by neural models, which are typically composed of an encoder and a decoder. Classic approaches typically employ heavy encoders with significantly more parameters than the decoder~\citep{kool2018attention,kwon2020pomo}. Recent studies, however, suggest that shifting parameters from the encoder to the decoder significantly enhances performance, particularly generalization ability~\citep{luo2023neural,drakulic2023bq,luo2025boosting,pirnay2024selfimprovement} (see Appendix~\ref{Related Work} for a detailed review).

Despite their success, these studies typically restrict the decoder size to 1–3M parameters, potentially underutilizing the decoder's capacity. No studies have yet investigated the effects of scaling the decoder size beyond this range, leaving a key question unanswered: \textit{how does performance evolve as the decoder scales up?} Addressing this question is crucial for NCO model development, as it provides guidelines for predicting performance gains and guiding efficient resource allocation.

To investigate this, we conduct preliminary tests across various parameter scales using a decoder-only architecture (to isolate the decoder's contribution). We train and test 12 models ranging from $\sim$1M to $\sim$150M parameters on TSP100, as detailed in Section~\ref{section: Experimental Setup}. Figure~\ref{fig:intro_efficient models} presents the results. While the optimality gap generally narrows as the number of parameters increases, the scaling behavior is non-monotonic. Notably, half of the models (6 of 12) outperform their larger counterparts (e.g., the 9.0M model outperforms the 41.5M model). These significant variances indicate that relying solely on parameter count cannot accurately predict the optimality gap. 
Instead, our experiments (see Section~\ref{section: Scaling Behavior} ) suggest that model depth (number of layers) and width (embedding dimension) are critical determinants.  
This observation raises a key question: \textit{What is the impact of depth and width on performance, and how can we leverage this to derive an efficient scaling strategy?}

To answer this question, this paper conducts a comprehensive study to quantify the scaling effects of depth and width. Specifically, we evaluate their impacts across three dimensions: parameter count, training data, and compute budget. Our empirical results identify a clear strategy: prioritizing depth over width consistently yields superior efficiency. Based on this insight, we establish practical design principles for the decoder-only model, guiding the efficient utilization of parameters, training data, and compute resources.

Our contributions can be summarized as follows: 
\begin{itemize} 
    \item We conduct the first systematic study on the scaling behavior of decoder-only models in NCO. We reveal that parameter count is not the sole determinant of performance; instead, model depth and width play critical roles in scaling efficiency. 
    \item We comprehensively evaluate the trade-offs between depth and width scaling across three dimensions: parameter, training data, and compute budget efficiency. Our empirical analysis demonstrates that deepening the model consistently yields superior performance gains compared to widening it.
    \item Building on these findings, we propose a set of practical design principles for the efficient utilization of parameters, compute budget, and training data. Extensive experiments validate the effectiveness of these principles. 
\end{itemize}

\section{Preliminary}

\subsection{Problem Definition}

We formulate the generic TSP as a problem on a fully connected graph $\mathcal{G} = (\mathcal{V}, \mathcal{E})$. Here, $\mathcal{V} = \{v_1, \dots, v_n\}$ represents a set of $n$ cities, where each node $v_i$ is associated with a coordinate vector $\mathbf{x}_i \in \mathbb{R}^d$ (typically $d=2$ for Euclidean TSP). The edge set $\mathcal{E}$ contains edges connecting all pairs of nodes, with weights $w_{ij} = \|\mathbf{x}_i - \mathbf{x}_j\|_2$ representing the Euclidean distance between city $v_i$ and $v_j$.The objective is to find a Hamiltonian cycle that visits every city exactly once and returns to the starting node, minimizing the total tour length. A solution is represented as a permutation $\boldsymbol{\pi} = (\pi_1, \dots, \pi_n)$ of the node indices $\{1, \dots, n\}$, where $\boldsymbol{\pi} \in \Omega$ denotes the space of all possible permutations. The optimization problem is formulated as:
\begin{equation}
    \boldsymbol{\pi}^* = \operatorname*{arg\,min}_{\boldsymbol{\pi} \in \Omega} \, \operatorname{Cost}(\boldsymbol{\pi})
    \label{eq:objective}
\end{equation}

\subsection{Construction-based Neural Routing Solver}

Construction-based solvers generate solutions $\boldsymbol{\pi}$ autoregressively, factorizing the joint probability as $p_\theta(\boldsymbol{\pi} | \mathcal{G}) = \prod_{t=1}^{n} p_\theta(\pi_t | \boldsymbol{\pi}_{<t}, \mathcal{G})$. The solver's architecture typically consists of two components: \textbf{(1) Encoder.} It maps raw inputs $\mathbf{x}_i$ to embeddings $\mathbf{h}_i \in \mathbb{R}^d$. In this work, to focus on the decoder-only architecture, we simplify the encoder to a single linear projection layer, projecting inputs directly to the embedding dimension $d$ without complex structural encoding. \textbf{(2) Decoder.} It functions as the next-node predictor. At step $t$, based on the currently visited sequence $\boldsymbol{\pi}_{<t}$, the decoder estimates a probability distribution over unvisited nodes to select the next city $\pi_t$.

\begin{table}[ht]
  \caption{Configuration of neural models with different depths and widths. ``Depth'' represents the number of layers in the decoder, and ``Width'' represents the dimension of the node embeddings. }
  \label{configuration of width and depth}
  \begin{center}
        \resizebox{0.99\columnwidth}{!}{
        \begin{tabular}{lccccr}
          \toprule
          \textbf{Depth}  & \textbf{Width} &  Head Num & QKV Dim  & FFN Dim & Parameters  \\
          \midrule
           & \textbf{128} & 8 & 16 & 512 &1.32M \\
          \textbf{6} & \textbf{256} & 16 & 16 & 1024&5.26M \\
           & \textbf{512} & 16 & 32 & 2048 &21.00M \\
          \midrule
           & \textbf{128} & 8 & 16 & 512 &2.60M \\
          \textbf{12} & \textbf{256} & 16 & 16 & 1024 &10.38M \\
           & \textbf{512} & 16 & 32 & 2048 &41.46M \\
          \midrule
           & \textbf{128} & 8 & 16 & 512 &5.17M \\
          \textbf{24} & \textbf{256} & 16 & 16 & 1024&20.62M \\
           & \textbf{512} & 16 & 32 & 2048&82.40M \\
          \midrule
           & \textbf{128} & 8 & 16 & 512 &9.02M \\
          \textbf{42} & \textbf{256} & 16 & 16 & 1024 &36.00M \\
           & \textbf{512} & 16 & 32 & 2048 &143.80M \\

          \bottomrule
        \end{tabular}
        }
  \end{center}
\end{table}

\begin{figure*}[t]
\centerline{\includegraphics[width=1.0\linewidth]{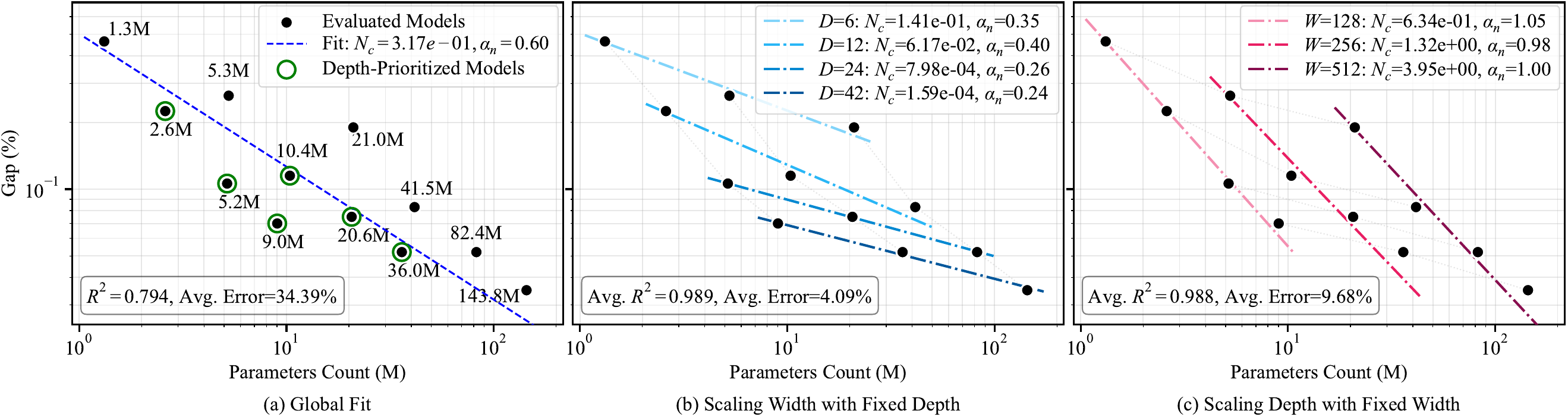}}
\caption{Scaling Law Analysis for Decoder-only Models. (a) Global Fit: Fitting a single power law (blue dashed line) to all models results in a poor fit ($R^2=0.794$) with a high average prediction error of $34.39\%$, indicating that parameter count alone is insufficient to predict performance. (b) Scaling Width vs. (c) Scaling Depth: We decouple the scaling behaviors by fixing one dimension. Comparing the scaling exponents ($\alpha_n$), we observe that scaling depth (c) yields significantly steeper slopes ($\alpha_n \approx 0.98 \text{--} 1.05$) compared to scaling width (b) ($\alpha_n \approx 0.24 \text{--} 0.40$). This quantitatively demonstrates that Depth-Prioritized Models exhibit a much faster rate of performance improvement than width-prioritized ones as parameters increase.}
\label{fig: 7 curves}
\end{figure*}

\section{Experimental Setup}
\label{section: Experimental Setup}

To provide a fine-grained analysis of different scaling strategies (depth vs. width), we train and evaluate a suite of 12 models with varying architectures in this section. 

\paragraph{Model Configuration}

Existing heavy decoder models typically have a minimum of 6 decoder layers and an embedding dimension of 128. Based on these, as shown in Table~\ref{configuration of width and depth}, we systematically increase the depths and widths by using the Cartesian product of four depths $L \in \{6, 12, 24, 42\}$ and three widths $d \in \{128, 256, 512\}$ to derive a comprehensive set of 12 models. This design space covers a diverse range of parameter counts from 1.32M to 143.8M and architectural depth-to-width ratios. The decoder-only model architecture is derived from existing heavy decoder-based models~\citep{luo2023neural,drakulic2023bq}, and adopts necessary architectural modifications detailed in Appendix~\ref{apl: model details}.

\paragraph{Training}
\label{Training}

Following existing works~\citep{luo2023neural,drakulic2023bq}, we train the model using supervised learning (SL) on a dataset of uniformly distributed 100-node TSP instances. We set the dataset size proportional to the total \textit{training steps} to ensure each instance is processed exactly once. This strategy prevents the dataset size from affecting model performance (see analysis in ~Appendix\ref{apl: training data}). To guarantee convergence across all models, we fix the total training steps at 60,000, with each step processing a batch size of 1024 instances. We initialize the learning rate at $1.25 \times 10^{-4}$ and decay it by a factor of 0.997 every 100 steps (see Appendix~\ref{apl: training details} for more training details).

\paragraph{Evaluation}

To evaluate model performance, we generate a test set of 10,000 uniformly distributed TSP100 instances, whose ground truth solutions are obtained using the LKH3 solver~\cite{LKH3}. We measure the model performance using the conventional metric Gap, which represents the percentage difference between the quality of the solution obtained by the model and the LKH3. We execute all evaluation experiments on a single NVIDIA RTX 4090 GPU with 24GB of memory. We refer to Appendix~\ref{In-domain performance} for the detailed results of the 12 evaluated models.

\section{Scaling Behaviors of Decoder-only Models with Different Depths and Widths }
\label{section: Scaling Behavior}

In this section, we first use a decoder-only Transformer to explore scaling behaviors across varying depths and widths. We examine these behaviors through three perspectives: parameter, training data, and compute efficiency. We then analyze the node selection process to determine how model shape influences these behaviors.

\subsection{Parameters Efficiency}
\label{section:scaling W&D}

Parameter efficiency measures the rate of gap reduction as the model size grows. To quantify it, we model performance as a function of parameter count using an empirical power law~\citep{kaplan2020scalinglawsneurallanguage}:
\begin{equation}
    \text{Gap}(N) =  (\frac{N_c}{N})^{\alpha_{n}},
    \label{eq:power law}
\end{equation}
where the exponent $\alpha_n$ quantifies scaling efficiency, that is, a larger $\alpha_n$ signifies higher parameter utilization, resulting in a steeper reduction of the optimality gap as the model size doubles, $N$ denotes the model parameter count with $\text{unit}=Millions$, and $N_c$ is a normalization coefficient. The quality of the fit is assessed using the coefficient of determination ($R^2$) and the mean absolute percentage error (denoted as Error). High $R^2$ values and low Error scores indicate that the empirical power law reliably describes the scaling behavior.

\paragraph{Insufficiency of Global Power-law Fit}
Figure~\ref{fig: 7 curves}(a) fits a unified power-law curve to the evaluation results of all 12 models. The regression yields a relatively loose correlation ($R^2 \approx 0.79$), failing to capture the fine-grained impact of architecture. As evidenced by the high average prediction error of 34.39\%, the data points exhibit significant dispersion around the fitted line. Specifically, depth-prioritized models consistently lie below the curve, while width-prioritized ones hover above it. This statistical divergence confirms that a coarse-grained law based solely on parameter count $N$ is insufficient for accurate performance prediction.

\paragraph{Scaling Efficiency: Depth vs. Width}

We attribute the global fitting failures to the distinct impacts of model depth and width. The efficient 9.0M model uses a depth-prioritized configuration ($D=42, W=128$), whereas the dominated models (21.0M and 41.5M) employ width-prioritized structures ($D \in \{6, 12\}, W=512$). To quantify this discrepancy, we isolate each dimension by holding one constant while varying the other, producing seven distinct scaling curves (Figure~\ref{fig: 7 curves}(b) and (c)).

Unlike the suboptimal global fit, decoupling the dimensions drastically reduces prediction error. As shown in Figure~\ref{fig: 7 curves}(b) and (c), the data points align tightly with the fitted power-law curves. Specifically, scaling width with fixed depth yields an average $R^2$ of 0.989 with a low prediction error of 4.09\%, while scaling depth with fixed width achieves an average $R^2$ of 0.988 and an error of 9.68\%. This order-of-magnitude improvement in goodness-of-fit (reducing error from $\sim34\%$ to $<10\%$) empirically validates that depth and width follow distinct scaling laws.
Crucially, the scaling exponents reveal a significant difference in efficiency: scaling depth yields exponents ($\alpha_n$) between 0.98 and 1.05, significantly outperforming scaling width, where $\alpha_n$ drops to between 0.24 and 0.40. This implies that depth scaling converts parameter growth into performance gains at a near-linear rate, i.e., doubling parameters reduces the gap to $\approx 50\%$ ($2^{-1.0}$) of its original value. In contrast, width scaling suffers from severe diminishing returns, reducing the gap to only $76\text{--}85\%$ ($2^{-0.24 \sim -0.40}$) of its original value. These findings establish that depth acts as the primary driver for performance, highlighting the inefficiency of purely widening the network.

\subsection{Data Efficiency}

\begin{figure}[t]
\centerline{\includegraphics[width=1.0\linewidth]{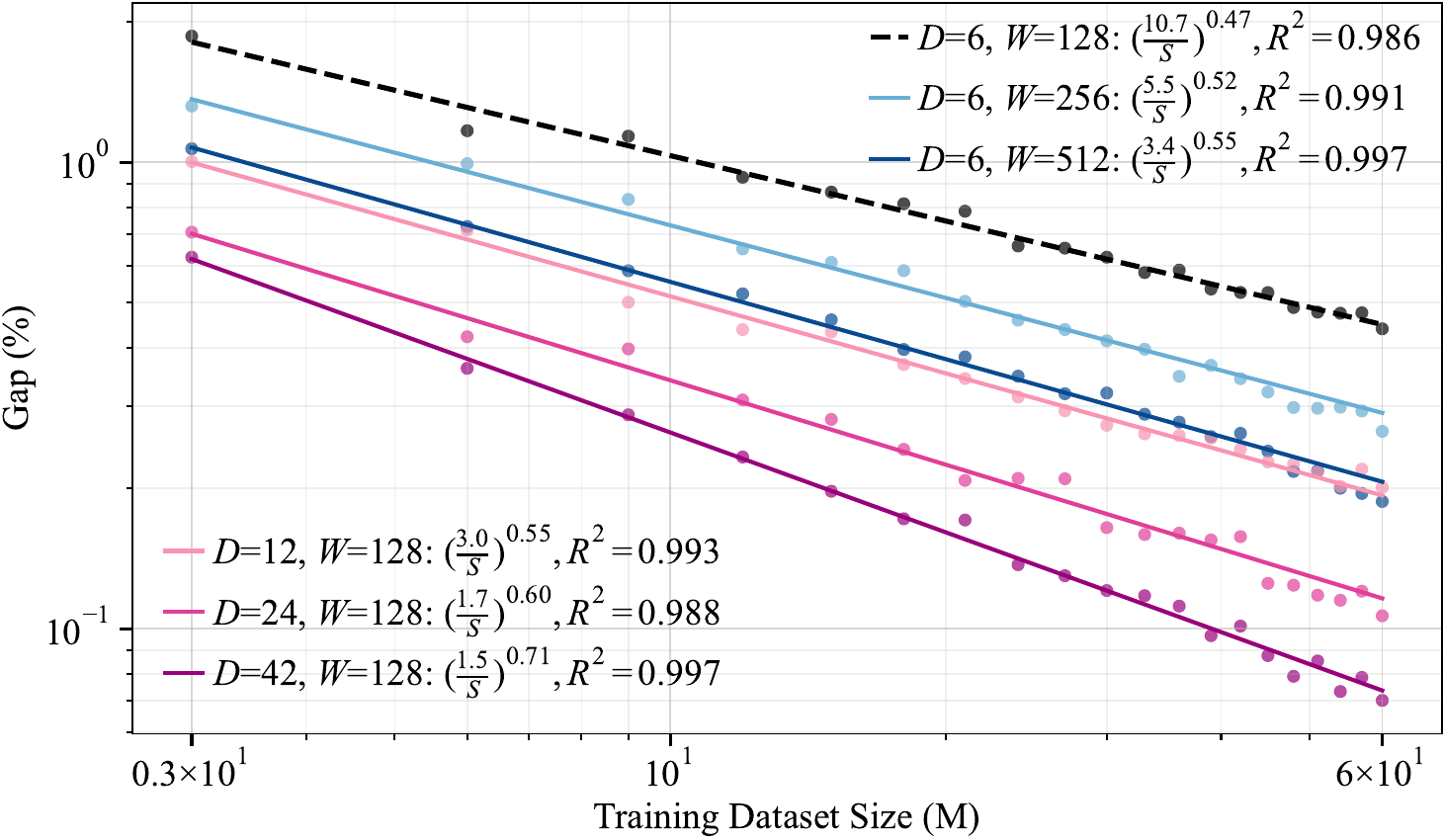}}
\caption{Data Efficiency Analysis. The log-log plot compares the optimality gap against training data size across different architectures. Prioritizing depth yields superior data efficiency: deeper models (e.g., $D=42$) exhibit steeper scaling exponents (up to $0.71$) compared to wider models (e.g., $W=512$, exponent $0.55$).
}
\label{fig: sample_efficiency_all}
\end{figure}

Beyond parameter efficiency, data efficiency, defined as the performance exhibited by a model given a fixed training dataset size, is another critical metric for evaluating scaling strategies. To quantify data efficiency, we compare deep models ($D \in \{12, 24, 42\}, W=128$) against wide variants ($D=6, W \in \{256, 512\}$). We track the progressive performance improvement on the TSP100 validation set as the training dataset scales from 3M to 60M samples, where each sample is seen exactly once.
For each model, we characterize this dynamic convergence trajectory using a power-law formulation, as shown in the equation below:
\begin{equation}
    Gap(S) = \left(\frac{S_c}{S}\right)^{\alpha_{s}},
    \label{eq: data_law}
\end{equation}
where $S$ represents the number of training samples, $S_c$ denotes the scaling constant, and $\alpha_{s}$ represents the data scaling exponent, which serves as a metric for the model's efficiency in fitting the training data. The visualized fitting results are illustrated in Figure~\ref{fig: sample_efficiency_all}.

The results reveal that scaling depth demonstrates exceptional data efficiency. Specifically, with the width fixed at $W=128$, we observe that the scaling exponent $\alpha_s$ increases monotonically with depth. For depths $D=6$ to 42, the scaling exponent $\alpha_s$ rises from 0.47 to 0.71, respectively. Meanwhile, the deep model ($D=42, W=128$) possesses significantly fewer parameters than the wide model ($D=6, W=512$), (9.02M vs. 21.00M), yet exhibits higher scaling efficiency $\alpha_s$ (0.71 vs. 0.55). This implies that doubling the training data reduces the optimality gap of the deep model to $61\%$ ($2^{-0.71}$) of its original value; by contrast, the wide model decays more slowly, reaching only $68\%$ ($2^{-0.55}$) of its original value. Furthermore, the power-law curves for the deeper models ($D=12/24/42; W=128$) consistently lie below those of the wider models ($D=6; W=256/512$). This indicates that deep-scaling models not only utilize parameters more effectively but also learn representation features much faster, resulting in superior data efficiency and significantly reduced data requirements.

\begin{figure*}[t]
\centerline{\includegraphics[width=0.85\linewidth]{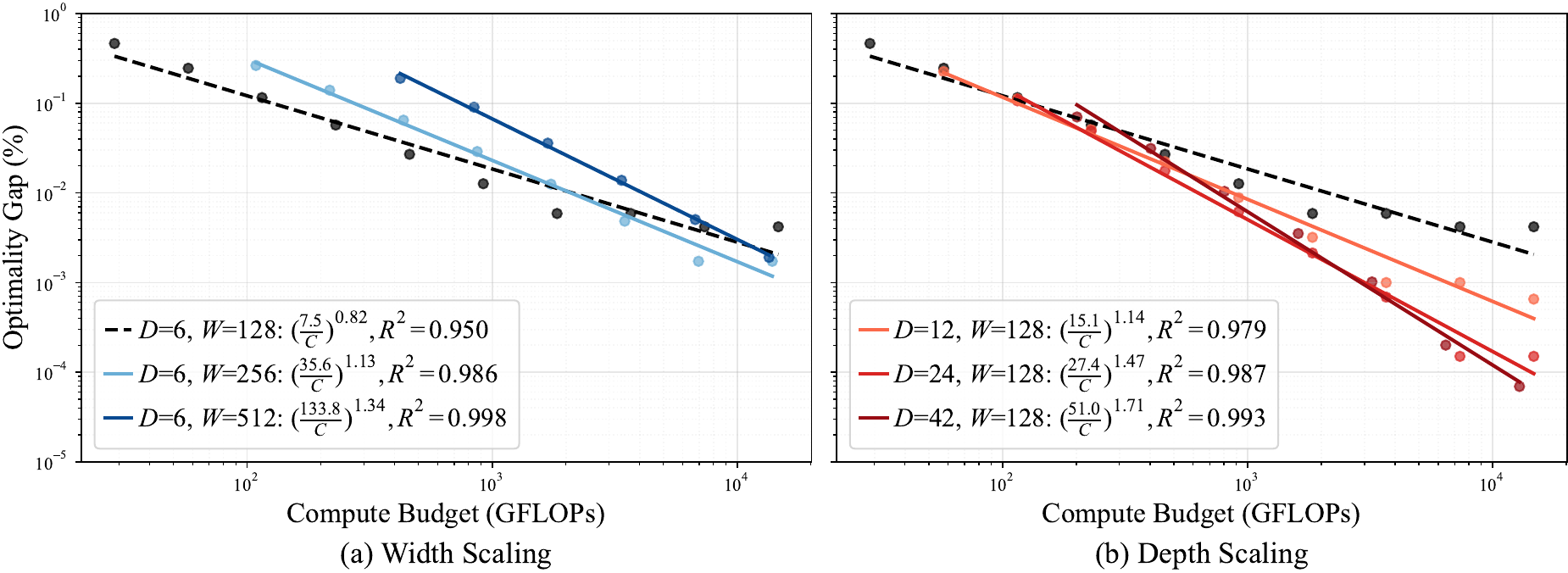}}
\caption{Compute Efficiency Analysis. Comparison of optimality gap vs. inference compute budget per instance (GFLOPs) for (a) width scaling and (b) depth scaling. Depth scaling demonstrates superior efficiency, achieving a significantly higher scaling exponent ($\alpha=1.71$ for $D=42$) compared to width scaling ($\alpha=1.34$ for $W=512$).}
\label{fig: computation efficiency}
\end{figure*}

\subsection{Compute Efficiency}

While increasing model parameters typically improves solution quality, it inevitably raises the compute budget. To identify effective compute budget scaling strategies, we quantitatively analyze the relationship between solution quality (optimality gap) and compute cost (FLOPs). We select FLOPs (floating-point operations) as a hardware-independent metric to eliminate the influence of hardware differences on inference time (refer to Appendix~\ref{Compute Budget and Inference Time} for real-world latency discussions).

We investigate the relationship between compute budget and the optimality gap by evaluating 12 model variants with different depths and widths on 10,000 TSP-100 instances. Specifically, we let the model perform inference using beam search with varying beam sizes to measure the resulting optimality gap and the average floating-point operations (FLOPs). These empirical data are fitted to the following power law:
\begin{equation}
\text{Gap}(C) = (\frac{C_c}{C})^{\alpha_c},
\label{eq:compute 1}
\end{equation}
where $C$ denotes the compute budget (in GFLOPs), and $C_c$ serves as a normalization coefficient. The exponent $\alpha_c$ quantifies the compute scaling efficiency, characterizing the rate at which the optimality gap decreases as the compute budget increases. The fitting results are shown in Figure~\ref{fig: computation efficiency}.

Specifically, Figure~\ref{fig: computation efficiency}(a) shows that at a fixed depth of 6, increasing model width yields moderate efficiency gains: as width expands from 128 to 512, the scaling exponent $\alpha_c$ rises from 0.82 to 1.34. This implies that doubling the compute budget reduces the gap to 56.6\% ($2^{-0.82}$) and 39.5\% ($2^{-1.34}$) of its original value, respectively. Conversely, Figure~\ref{fig: computation efficiency}(b) reveals that increasing depth at a fixed width of 128 delivers a more substantial efficiency boost. As depth increases from 6 to 42, $\alpha_c$ improves sharply from 0.82 to 1.71, meaning a $2\times$ budget increase reduces the gap to 30.6\% ($2^{-1.71}$) of its original value. This efficiency difference is clear under a fixed compute budget $\approx 10^4$ GFLOPs, where the deepest model ($D=42$) achieves an optimality gap orders of magnitude lower than the widest baseline ($W=512$). Consequently, driven by a significantly steeper decay rate, deep scaling proves to be a more effective strategy than width scaling for minimizing the optimality gap under fixed compute budgets.

\section{Design Principles}

Synthesizing our findings on depth and width scaling behaviors, we distill a set of design principles to guide the development of scalable models. These principles are framed around three key dimensions: model parameters, training data, and compute budget.

\begin{figure}[t]
\centerline{\includegraphics[width=0.9\linewidth]{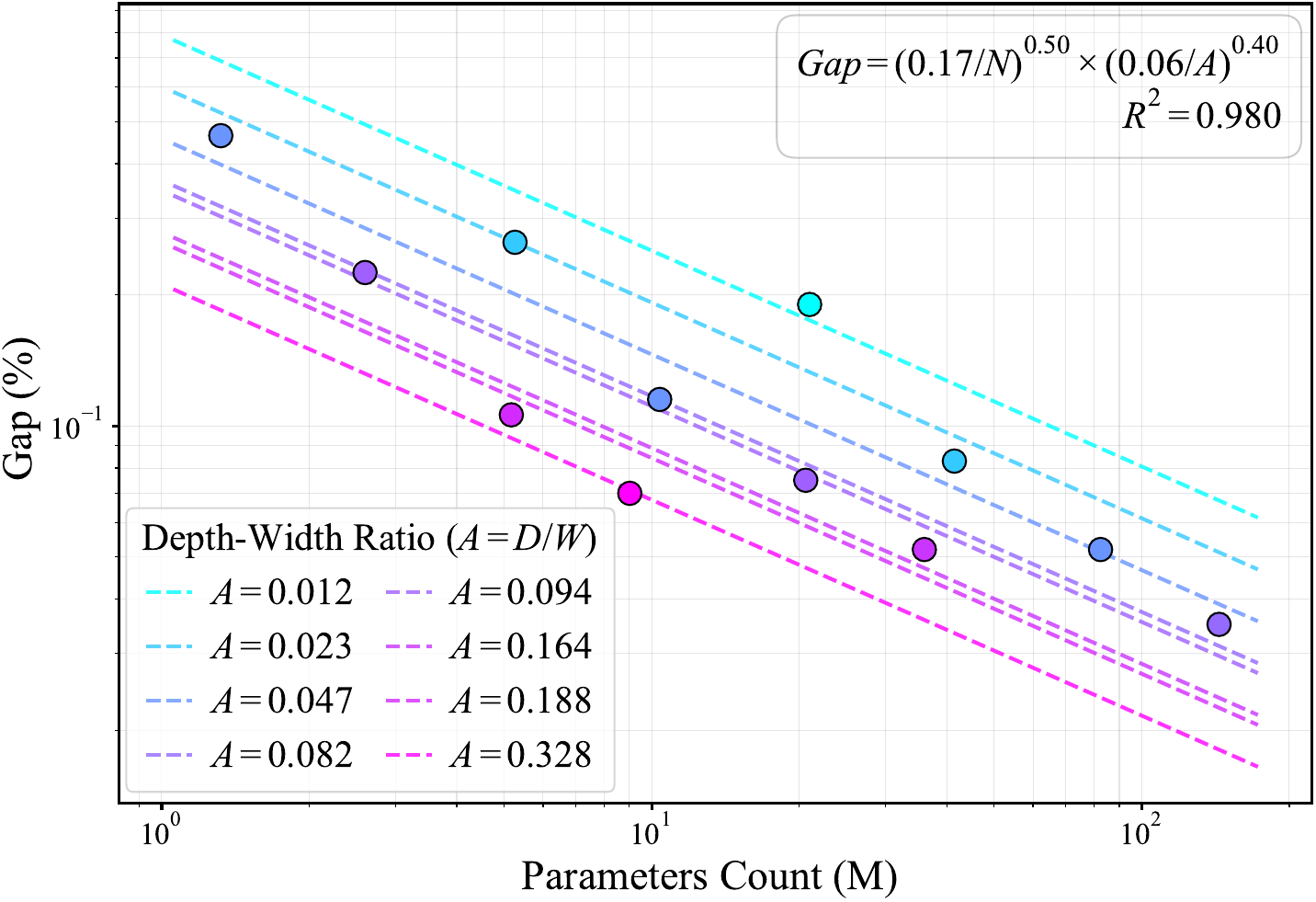}}
\caption{Visualization of the performance gap against model parameters ($N$) in log scale for varying depth-to-width ratios ($A$). The consistent drop in the gap as lines shift from cyan (low $A$) to pink (high $A$) illustrates that prioritizing deeper, narrower architectures effectively reduces the optimality gap under constrained parameter budgets.}
\label{fig: n_and_r}
\end{figure}

\subsection{Parameter Placement Policy}
\label{Parameter Placement Policy}
Building on the finding that deeper models are more parameter-efficient, we turn to the practical question of how to allocate a given parameter budget. This requires understanding the quantitative interplay between depth and width. 
Extending Equation~\ref{eq:power law}, we model the performance gap in terms of model width $W$ and depth $D$ using a bivariate power-law function: 
\begin{equation} \text{Gap}(W,D) = \left(\frac{D_c}{D}\right)^{\beta_d} \left(\frac{W_c}{W}\right)^{\beta_w}.
\label{eq:gap_wd} \end{equation}
Then, by substituting the depth-to-width ratio $A = D/W$ and the parameter approximation $N \approx cDW^2$ into this bivariate power-law function, we formulate the Gap directly in terms of the parameter count $N$ and the model shape $A$:
\begin{equation}
\text{Gap}(N,A) = \left(\frac{N_c}{N}\right)^{\beta_{n}} \left(\frac{A_c}{A}\right)^{\beta_{a}}.
\label{eq:gap_nr}
\end{equation}
Using the evaluation results of the 12 models for fitting, we obtain the coefficient values of Equation~\ref{eq:gap_nr}, and visualize the results to clarify the joint impact of parameter count and depth-to-width ratio on performance in Figure~\ref {fig: n_and_r}. The results reveal that architectural shape ($A$) is nearly as critical as model size ($N$). Specifically, doubling the parameter count $N$ (while fixing $A$) reduces the gap to 71\% ($2^{-0.5}$) of its original value, while doubling the depth-to-width ratio $A$ (while fixing $N$) yields a comparable benefit, reducing the gap to 76\% ($2^{-0.4}$). These findings highlight that geometric scaling is as vital as parameter scaling, leading to the following principle:

\textbf{Design Principle:} \textit{To maximize performance, models should adopt a deep-narrow architecture where depth is prioritized over width.}

\textbf{Validation} We empirically verify this principle by optimizing the decoder-only model with the configuration $D=6, W=128$ under a fixed parameter budget. We restructure the original configuration into deeper, narrower variants ($D=24, W=64$) and ($D=96, W=32$). We evaluate these models on 10,000 TSP100 instances and 128 instances for each scale from TSP200 to TSP1000. The results are detailed in Table~\ref{tab: principle 1}.
These results demonstrate that architectural adjustment driven by our design principle yields substantial improvements. Despite using fewer parameters, the optimized deep-narrow model ($D=96, W=32$) not only enhances in-domain performance from 0.464\% to 0.184\% (reducing the TSP100 gap by $60.3\%$) but also demonstrates significantly stronger out-of-domain generalization across TSP200--1000. Most notably, on the large-scale TSP1000, the optimality gap is reduced from $2.172\%$ to $0.991\%$, corresponding to a substantial $54.4\%$ relative improvement. These results provide robust empirical verification of our design principle.

\begin{table}[h]
\centering
\caption{Validation of Design Principle: Performance comparison of model variants with varying depth-to-width ratios under a fixed parameter budget ($\approx 1.3$M)} 
\resizebox{1.0\linewidth}{!}{%
\label{tab: principle 1}
\begin{tabular}{l|c|c|c|c|c}
\toprule
Model & Parameters & TSP100 & TSP200 & TSP500 & TSP1000 \\
\midrule
$D=6,W=128$ & 1.32M & 0.464\% & 0.669\% & 1.559\% & 2.172\% \\
$D=24,W=64$ & 1.29M &  0.297\% & 0.449\% & 0.857\% & 1.259\% \\
$D=96,W=32$ & 1.30M & \textbf{0.184\%} & \textbf{0.307\%} & \textbf{0.678\%} & \textbf{0.991\%} \\
\bottomrule
\end{tabular}
}
\end{table}

\subsection{Data Efficiency Policy}

Our findings reveal that scaling depth demonstrates exceptional data efficiency. Specifically, deeper models exhibit a steeper power-law exponent ($\alpha_s$), implying they can capture complex patterns from limited data more effectively. This leads to the following principle:

\textbf{Design Principle:} \textit{To optimize data efficiency, particularly in data-scarce regimes, we should prioritize model depth, as deeper architectures utilize limited training samples more effectively.}

\textbf{Validation:} 
To validate the efficacy of this principle within a data-constrained regime, we limit the training dataset size to 0.1M samples. On this fixed dataset, we employ a multi-epoch training strategy, cycling through the samples for a budget of 60 epochs. We compare three distinct architectures: a baseline ($D=6, W=128$), a deep counterpart ($D=42, W=128$), and a wide counterpart ($D=6, W=512$). We evaluate their performance across 10,000 TSP100 and 128 TSP1000 instances, with the results summarized in Table~\ref{tb:sample utilization policy}.

The results demonstrate the advantage of depth scaling under dataset size constraints. First, simply expanding the model width yields diminishing returns: the wide model ($D=6, W=512$) reduces the optimality gap from the baseline's $1.715\%$ to $0.949\%$, but this improvement demands a massive parameter increase of nearly $16\times$ (from 1.32M to 21.0M). In contrast, increasing depth is far more efficient. The deep configuration ($D=42, W=128$) achieves a significantly superior gap of $0.547\%$, which is nearly half that of the wide model ($0.949\%$) while utilizing less than half the parameters (9.02M vs. 21.0M). Furthermore, this advantage extends to out-of-domain generalization: on large-scale TSP1000 instances, the deep model reduces the gap to 2.572\%, significantly outperforming the wide model's 3.871\%. This indicates that the deep architecture extracts more robust representations per parameter from the limited data budget. These empirical findings strongly validate the effectiveness of our proposed design principle regarding data efficiency.

\begin{table}[ht]
  \caption{Validation of Data Efficiency Policy: Performance comparison of depth versus width scaling under a limited training budget ($10^5$ samples).}
  \label{tb:sample utilization policy}
  \begin{center}
    \begin{small}
      \resizebox{1.0\columnwidth}{!}{
        \begin{tabular}{cccc}
          \toprule
            Model (Parameters)  & Training Data Size & TSP100 & TSP1000 \\
          \midrule
          $D=6,W=128$ (1.32M) & 0.1M & 1.715\% & 5.678\% \\
          $D=6,W=512$ (21.0M) & 0.1M & 0.949\% & 3.871\% \\
          $D=42,W=128$ (9.02M) & 0.1M & \textbf{0.547\%} & \textbf{2.572\%} \\
          \bottomrule
        \end{tabular}
        }
    \end{small}
  \end{center}
\end{table}

\subsection{Computation Allocation Policy}

\begin{figure}[t]
\centerline{\includegraphics[width=1.0\linewidth]{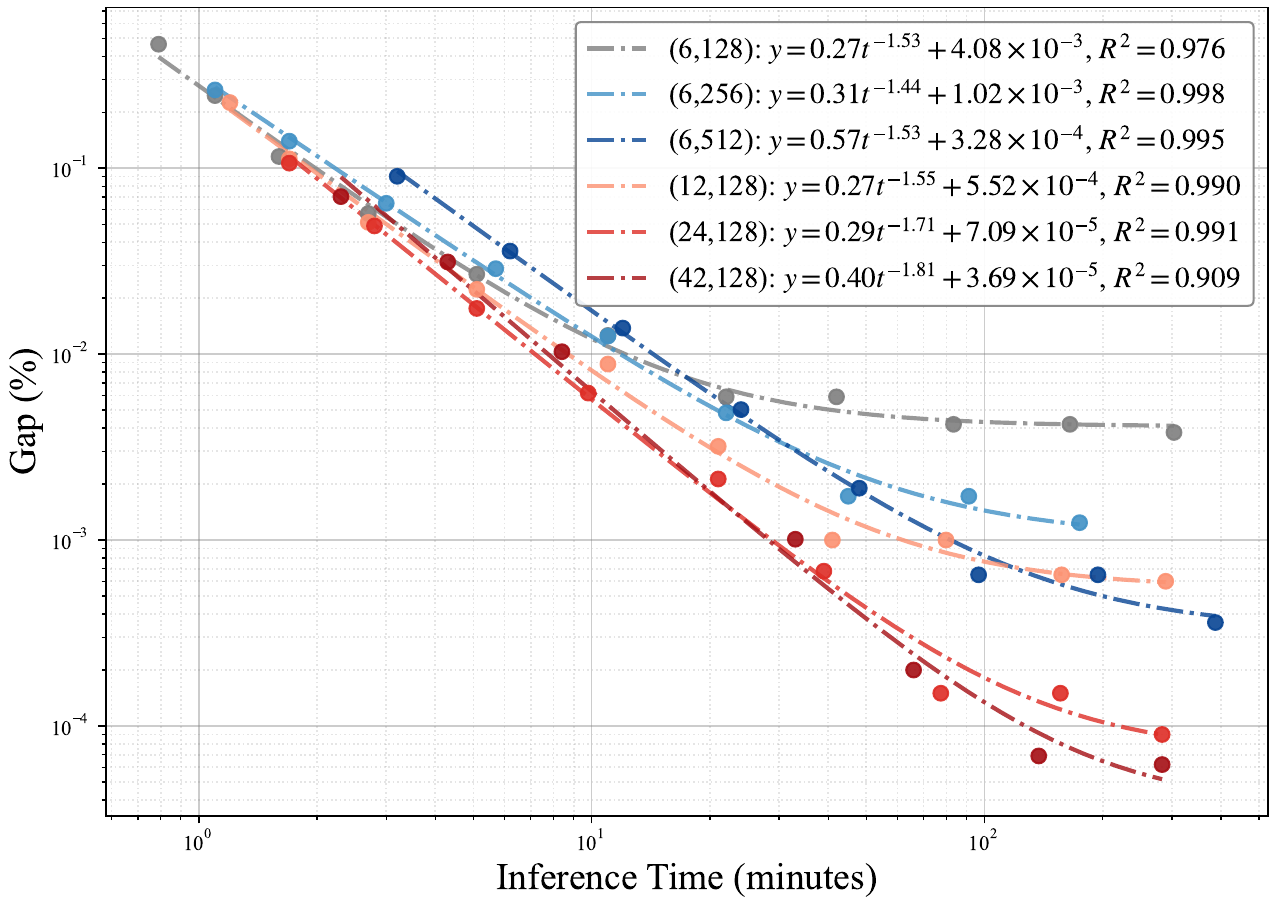}}
\caption{ Scaling laws of inference performance across different model configurations. The plot depicts the relationship between optimality gap and inference time (minutes) on a log-log scale. Dashed lines indicate best-fit curves using a shifted power-law model ($y = \alpha_t t^{-\beta_t} + \gamma$), where $t$ denotes inference time. Here, $\alpha_t$ serves as the scaling coefficient, and $\gamma$ represents the asymptotic lower bound of the gap as $t \to \infty$. The parameter $\beta_t$ denotes the decay exponent, indicating the magnitude of performance gain for every doubling of the inference budget.}
\label{fig: beam_time}
\end{figure}

Our theoretical analysis indicates that depth scaling yields superior compute efficiency, achieving higher solution quality than width scaling under equivalent budgets. To translate these theoretical metrics into practical deployment constraints, we analyze the optimality gap as a function of inference time. Specifically, we implement test-time scaling by increasing the beam width and fit the resulting experimental data to the following power-law formula:
\begin{equation}
    \text{Gap}(t) = \alpha_{t} t ^ {-\beta_{t}} + \gamma, 
    \label{eq: beam and time}
\end{equation}
The visualization results in Figure~\ref{fig: beam_time} corroborate our theoretical analysis. Building on this consistency, we formalize our findings into the following design principle:

\textbf{Design Principle:} \textit{To maximize compute efficiency, we should prioritize using medium-depth models under medium inference budgets; in the case where inference budgets are sufficient, use deeper models instead.}

\textbf{Validation:} To investigate the inference efficiency of different scaling strategies, we evaluate model performance on TSP100 with 1,000 instances under varying inference budgets controlled by beam search width. We compare four distinct architectures: a baseline ($D=6, W=128$), a wide variant ($D=6, W=512$), and two deep variants ($D=24, W=128$ and $D=42, W=128$). We analyze their behavior across two resource regimes: (1) Limited Inference Budget: 
We restrict the wall-clock time to a fixed window ($\approx 12$ minutes). We adjust the beam size for each model to strictly adhere to this limit and compare the resulting solution quality. (2) Sufficient Inference Budget: To determine the performance upper bounds, we provide a sufficient compute budget ($> 1$ hour). This allows us to scale the beam width extensively, observing the asymptotic lower bound of the optimality gap for each architecture. The results are shown in Table~\ref{tb:inference_budget}.

The results demonstrate that under latency constraints ($\approx 12$ minutes), the medium-depth model ($D=24, W=128$) proves significantly more cost-effective than width scaling, outperforming the wide variant while utilizing substantially fewer parameters (5.2M vs. 21.0M), as shown in Table~\ref{tb:inference_budget}. It achieves a superior optimality gap of 0.0056\% compared to the wide model's 0.0138\% ($D=6, W=512$). This confirms that in resource-constrained regimes, a medium-depth model yields a more favorable performance-latency trade-off than increasing width. When the inference budget is sufficient ($> 1$ hour), the deep model ($D=42, W=128$) targets the highest performance ceiling, achieving a near-zero gap of 0.0001\%. Crucially, it proves not only more accurate but also more time-efficient than the medium-depth model ($D=24, W=128$). The $D=42$ model reaches this superior result in 2.29 hours, whereas the $D=24$ model requires more time (2.60 hours) to converge to a suboptimal gap of 0.0002\%. This confirms that depth scaling allows the model to push the boundaries of extreme optimality.

\begin{table}[t]
    \centering
    \caption{Performance comparison under different inference budgets. We report the Optimality Gap (\%) and total inference time. `bs' denotes Beam Search.}
    \label{tb:inference_budget}
    \resizebox{\linewidth}{!}{
        \begin{tabular}{l ccc ccc}
            \toprule
            \multirow{2}{*}{(Depth,Width,Params)} & \multicolumn{3}{c}{Limited Budget} & \multicolumn{3}{c}{Sufficient Budget ($>$1h)} \\
            \cmidrule(lr){2-4} \cmidrule(lr){5-7}
            & bs & Gap (\%) & Time & bs & Gap (\%) & Time \\
            \midrule
            (6, 128, 1.3M)  & 32 & 0.0127 & 12m & 512 & 0.0059 & 2.75h \\
            (6, 512, 21.0M) & 8  & 0.0138 & 12m & 128 & 0.0007 & 3.24h \\
            (24, 128, 5.2M) & 10 & \textbf{0.0044} & 12m & 128 & 0.0002 & 2.60h \\
            (42, 128, 9.0M) & 6  & 0.0056 & 12m & 64  & \textbf{0.0001} & 2.29h \\
            \bottomrule
        \end{tabular}
    }
\end{table}

\begin{table}[t]
\centering
\caption{Performance of recent advanced NCO constructive methods on TSP instances with different problem sizes and data distribution. Ground truth solutions are obtained via LKH3~\cite{LKH3} for uniform datasets. `bs' denotes Beam Search.}

\label{tb:result-out-of-domain}
\resizebox{0.83\columnwidth}{!}{%
\begin{tabular}{l|c c|c c}
\toprule[0.5mm]
 & \multicolumn{2}{c|}{TSP 100} &  \multicolumn{2}{c}{TSP 1000}  \\
 & Gap & Time & Gap & Time \\
 \midrule
 LEHD greedy  & 0.577\% & 0.29m &  3.168\% & 1.17m \\
 LEHD bs16 & 0.038\% & 4.4m & 1.874\% & 22m  \\
 BQ greedy  & 0.348\% & 1.18m & 2.274\% & 1.18m  \\
 BQ bs16  & 0.015\% & 17.6m & 1.35\% & 18m  \\
 SIGD greedy & 0.851\% & 2.4m &  2.791\% & 2.71m  \\
 SIGD bs16 & 0.092\% & 28m &  1.569\% & 39m \\
 \midrule
 Ours (Depth, Width)  \\
 Ours (6, 128) greedy & 0.464\% & 0.34m &  2.172\% & 1.6m  \\
 Ours (6, 128) bs2 & 0.246\% & 1.1m &  1.819\% & 3.8m \\
 Ours (6, 512) greedy & 0.190\% & 1.7m  & 1.291\% & 4.2m  \\
 Ours (6, 512) bs2 & 0.090\% & 3.2m  & 1.055\% & 8.7m \\
 Ours (42, 128) greedy & 0.070\% & 2.0m & 0.869\% & 11m  \\
  Ours (42, 128) bs2 & 0.031\% & 4.3m & 0.735\% & 22m  \\
  Ours (42, 512) greedy & 0.035\% & 11m  & 0.576\% & 28m  \\
 Ours (42, 512) bs2& \textbf{0.013\%} & 21m  & \textbf{0.493\%} & 57m  \\
\bottomrule[0.5mm]
\end{tabular}%
}
\end{table}

\begin{table}[t]
    \centering
    \caption{Performance comparison (Gap \%) between baselines and our models in TSPLib. Ground truth solutions are obtained from TSPLib~\citep{reinelt1991tsplib}.}
    \label{tab:performance_comparison lib}
    \resizebox{\columnwidth}{!}{
        \begin{tabular}{l ccc cccc}
            \toprule
            \multirow{2}{*}{TSPLib} & \multicolumn{3}{c}{Baselines} & \multicolumn{4}{c}{Ours (Depth,Width)} \\
            \cmidrule(lr){2-4} \cmidrule(lr){5-8}
            & BQ & SIGD & LEHD & (6,128) & (6,512) & (42,128) & (42,512) \\
            \midrule
            $<500$ & 2.828\% & 2.176\%  & 2.287\%  & 2.582\%  & 1.553\%  & 1.364\%  & \textbf{0.811\% } \\
            $>500$ & 10.288\%  & 10.812\%  & 9.720\%  & 7.249\%  & 4.127\%  & 3.631\%  & \textbf{1.942\% } \\
            \midrule
            All & 5.812\%  & 5.631\%  & 5.260\%  & 5.235\%  & 3.108\%  & 2.346\%  & \textbf{1.288\% } \\
            \bottomrule
        \end{tabular}
    }
\end{table}

\section{Validation of the Scaling Principle on Generalization}

We extend to out-of-distribution scenarios to validate the effectiveness of the design principles. We select four representative configurations from the design space introduced in Section~\ref{section: Experimental Setup}. These models represent the boundaries of different scaling strategies: (1) The baseline ($D=6, W=128$); (2) The wide variant ($D=6, W=512$); (3) The deep variant ($D=42, W=128$); (4) The fully scaled model ($D=42, W=512$). This selection allows us to rigorously isolate the impact of depth versus width on generalization.

Given that our derived scaling principles are closely tied to sequential decision-making processes, we limit our scope to autoregressive construction-based methods, with a particular emphasis on heavy-decoder architectures and constructive decoding strategies. Consequently, we benchmark our models against state-of-the-art methods within this category, including LEHD~\citep{luo2023neural}, BQ~\citep{drakulic2023bq}, SIGD~\citep{pirnay2024selfimprovement}. All models are trained on Uniform-TSP100 instances. We evaluate performance on two distinct test datasets: (1) uniformly distributed synthetic datasets with 100 and 1,000 nodes (10K and 128 instances, respectively), and (2) TSPLIB datasets ($N < 5,000$) to verify performance on real-world distributions. We report both the optimality gap and total inference latency using two construction-based decoding strategies: greedy search and beam search. The results are shown in Tables~\ref{tb:result-out-of-domain} and~\ref{tab:performance_comparison lib}.

From these results, the deep architecture exhibits exceptional generalization on larger instances. On TSP1000, the $D=42, W=128$ model (Greedy) significantly outperforms all baseline methods equipped with extensive beam search ($B=16$). It achieves superior solution quality (0.869\%) compared to the strongest baseline (BQ~\citep{drakulic2023bq}: 1.35\%) while requiring roughly half the inference time (11m vs. 18m). 
Comparing with the wide variant ($D=6, W=512$), the deep variant ($D=42, W=128$) yields significantly lower gaps on TSP1000 (0.869\% vs. 1.291\%) and TSPLib (2.346\% vs. 3.108\%) in greedy decoding, even with a smaller parameter budget. This reveals that prioritizing depth over width greatly enhances cross-scale generalization. When comparing models with identical widths (e.g., $D=42$ vs. $D=6$ at $W=512$), the deeper models consistently achieve lower gaps.
Finally, building upon this depth-prioritized foundation, our fully scaled model ($D=42, W=512$) leverages its increased capacity to establish a new state-of-the-art, reducing the optimality gap by 73.4\% on TSP1000 compared to the baseline in greedy decoding. Notably, both deep models ($D=42$) break the 1\% error barrier on TSP1000. To the best of our knowledge, this marks the first time that end-to-end constructive NCO models have achieved an optimality gap lower than 1\% on TSP1000 using purely greedy decoding. Appendices~\ref{app: Inference using Beam Search} and~\ref{app: Inference using RRC} show further performance gains using more compute budget.

\vspace{-6pt} 
\section{Additional Studies}
\vspace{-2pt} 

\paragraph{Representation Quality and Long-sightedness} Analysis in Appendix~\ref{Additional Analysis: Scaling Depth vs. Width} using PCA, cosine similarity, and a novel long-sightedness metric reveals that deep scaling fosters a more structured embedding space, significantly enhancing the distinction of optimal nodes and distant targets.
\vspace{-7pt} 

\paragraph{Versatility of Design Principle} Experiments on the LEHD baseline (Appendix~\ref{Versatility of Design Principle 1}) confirm that our deep-narrow principle transfers to other architectures, achieving substantial performance gains on both in-domain and out-of-domain instances with fewer parameters.

\vspace{-9pt} 
\section{Conclusion, Limitation, and Future Work}
\vspace{-2pt} 

In this work, we investigate strategies for efficiently scaling the NCO model's Decoder. By evaluating efficiency across three dimensions: parameters, training data, and computation, we discover that increasing depth consistently outperforms increasing width. Based on this analysis, we propose three design principles: (1) deep-narrow architectures generally are superior; (2) deep models maximize efficiency when training data is scarce; and (3) model depth should scale with the compute budget, favoring medium depth for constrained resources and greater depth for abundant ones. Empirical validation demonstrates the effectiveness of scaling the decoder using the depth-first principles, showing that the resulting models achieve notable improvements on both in-domain and out-of-domain performance.

\textbf{Limitation and Future Work} 
Our experiments focus on the parameter range from 1M to 144M. While models of this scale already exhibit clear trends, investigating the scaling behaviors at a larger scale is a promising future work. In addition, while this paper focuses on the fundamental TSP to isolate and characterize scaling behaviors, a promising future work is to study the scaling analysis on other routing problems, such as the capacitated vehicle routing problem.


\section*{Impact Statement}
This paper presents work whose goal is to advance the field of Machine Learning. There are many potential societal consequences of our work, none of which we feel must be specifically highlighted here.

\bibliography{example_paper}
\bibliographystyle{icml2026}

\newpage
\appendix
\onecolumn

\section{Related Work}
\label{Related Work}

\subsection{Construction-based Methods}

Autoregressive models generate solutions sequentially. In this paradigm, a neural model iteratively predicts the probability of adding an unvisited node to a partial solution. \citet{vinyals2015pointer} establishes this approach by training Pointer Networks (Ptr-Nets) via Supervised Learning (SL). Subsequent works~\citep{bello2016neural,nazari2018reinforcement} demonstrate that Reinforcement Learning (RL) can further improve performance on small-scale problems. Leveraging the Transformer architecture~\cite{vaswani2017attention}, \citet{kool2018attention} introduces the Attention Model (AM), which achieves outstanding results on Vehicle Routing Problems (VRPs) with up to 100 nodes. To bridge the gap between neural approaches and traditional heuristics, researchers have developed various AM-based variants through improved architectural designs, training strategies, and inference techniques. These variants generally fall into two categories: heavy-encoder-and-light-decoder (HELD)~\citep{khalil2017learning,kwon2020pomo,xin2020step,xin2021multi,hottung2021efficient,kim2022sym,choo2022simulation,bi2022learning,jin2023pointerformer,xiao2023distilling,gao2023towards,kim2021learning,bi2024learning,sun2024learning,zheng2024dpn,fang2024invit,huang2025rethinking,hottung2025polynet,weissman2025complexity,zhou2024icam,zhou2025urs} and light-encoder-and-heavy-decoder (LEHD)~\citep{luo2023neural,drakulic2023bq,luo2025boosting,drakulic2025goal,luo2025rethink_tightness,luo2025learning,pirnay2024selfimprovement,zhou2025l2rlearningreducesearch,drakulic2025goal,ijcai2025DGL,chen2025improving,zheng2025mtlkd}. While heavy encoder-based methods show promising results, heavy decoder-based methods has greater potential narrow the gap especially on large-scale problems. Nevertheless, heavy decoder-based methods restrict the decoder's parameters to a relatively small scale, such as 1.43M for LEHD~\citep{luo2023neural} and 3.11M for BQ~\citep{drakulic2023bq}. Intuitively, increasing parameters enhances performance using a larger model capacity. However, no studies have been conducted to investigate the effects of scaling beyond this range. In addition, scaling the model parameter needs to consider two factors: the model's depth (the number of layers) and width (the size of embedding). Effective scaling requires a precise understanding of these two factors. Consequently, this paper investigates how model depth and width influence performance, aiming to establish design guidelines for effective parameter scaling. To bridge this gap, we investigate two scaling strategies: scaling depth versus scaling width. We conduct a comprehensive analysis in three aspects of parameter, training data, and compute efficiency to identify the superior approach. Based on our findings, we provide design guidelines for scaling decoders tailored to various practical applications.

Non-autoregressive (NAR) methods typically formulate the problem as an edge prediction task. These approaches utilize Graph Neural Networks (GNNs) to generate an edge probability heatmap, from which complete solutions are constructed~\citep{joshi2019efficient,fu2021generalize,qiu2022dimes,li2023from,min2023unsupervised,sun2023difusco}. Typically, achieving high-quality solutions often necessitates sophisticated post-search strategies, such as Monte Carlo Tree Search (MCTS)~\citep{fu2021generalize}. Furthermore, extending these heatmap-based methods to complex Vehicle Routing Problem (VRP) variants proves challenging due to intricate constraint handling. Consequently, this paper focuses on the autoregressive approach.

\subsection{Improvement-based Methods}

In contrast to construction-based approaches that generate solutions node-by-node, improvement-based methods refine complete solutions. Starting with an initial solution, these methods iteratively apply local modifications to enhance quality. Literature typically categorizes these approaches into Large Neighborhood Search (LNS) and Small Neighborhood Search. LNS methods generally adopt a `divide-and-conquer' paradigm, decomposing large-scale problems into tractable sub-problems to be solved independently and then merged~\citep{li2021learning,zong2022rbg,hou2023generalize,pan2023h-tsp,ye2024glop,zheng2024udc}. Within this framework, the sub-problem solver is typically trained to generate either complete solutions for small-scale instances or partial solution segments (e.g., sub-tours in TSP) for larger graphs~\citep{kim2021learning,cheng2023select,luo2023neural}. Conversely, Small Neighborhood Search methods utilize operators such as 2-opt and swap to perturb and recombine local solution segments~\citep{chen2019learning,lu2019learning,d2020learning,paulo2020learning,ma2021learning,wu2021learning,ma2023neuopt}. Due to the limited scope of these perturbations, such methods are primarily effective for small-scale problems. Beyond these categories, alternative non-constructive NCO approaches include augmenting classical solvers with neural components~\citep{xin2021neurolkh} or solving VRPs at a route level~\citep{delarue2020reinforcement}. Distinct from these improvement-based paradigms, this paper explores how to efficiently scale the decoder within the framework of autoregressive construction-based methods.

\newpage

\section{Model Structure}
\label{apl: model details}

To obtain our neural routing solver, we use the model architecture of LEHD~\citep{luo2023neural} but introduce several refinements. First,  to eliminate the influence of the encoder as we mainly focus on the scaling strategy for the decoder, we remove the encoder. Furthermore, to enhance the stability of the training as the decoder scales up, we incorporate the Gated Attention~\cite{qiu2025gatedattentionlargelanguage} and Rezero Normalization~\citep{bachlechner2021rezero} into the model following~\citep{drakulic2023bq}. The model mainly consists of an embedding layer, a multiple-layer decoder, and an output layer. The details are described below.

\subsection{Embedding Layer}

Given TSP instances, we linearly project the 2D coordinates of all nodes ($\mathbf{n}_{1}$,$\mathbf{n}_{2}$...$\mathbf{n}_{N}$) into a high-dimensional embedding space to obtain initial node representations($\mathbf{h}_{1}$,$\mathbf{h}_{2}$...$\mathbf{h}_{N}$). At step $t$, the starting and current nodes are processed via distinct linear transformations to encode trajectory context ($\mathbf{h'}_{1}$,$\mathbf{h'}_{t}$). The embeddings of available nodes are represented by $\mathbf{h}_{available}$. Finally, we concatenate the embeddings of the start node embedding, current node embedding , and available nodes embeddings to form the input $X$ for the Decoder Layer in Equation~\ref{eq: initial embedding}.
\begin{equation}
     X = \operatorname{concat}({\mathbf{h'}}_{1},\mathbf{h}_{available},\mathbf{h'}_{t})
    \label{eq: initial embedding}
\end{equation}

\subsection{Decoder Layer}

In the decoder layer, we incorporate a gating mechanism into the attention layer, following~\cite{qiu2025gatedattentionlargelanguage}. Additionally, we employ ReZero normalization to improve the residual connections. The specific implementation details are as follows.

\textbf{Gated Attention}
The Gated Attention Layer processes the input embedding matrix $X$ through feature aggregation to produce the output embedding $H$. 
This process comprises two parallel branches. In the first branch, the input $X$ is linearly projected into queries $Q$, keys $K$, and values $V$ using weight matrices $W_Q, W_K,$ and $W_V$ as shown in Equation~\ref{eq:Q,K,V}. We then compute the mutlti-head attention output $\text{Attention}(XW_Q^{(i)}, XW_K^{(i)}, XW_V^{(i)})$ based on these components. 
In parallel, as shown in Equation~\ref{eq: gated G}, the gating signal $G$ is derived from $X$ via a separate transformation in the second branch. 
Finally, the output $H$ is obtained by performing element-wise multiplication between the attention output and the gate $G$ as shown in Equation~\ref{eq: output H}.
\begin{equation}
\begin{aligned}
    \text{head}_i &= \text{Attention}(XW_Q^{(i)}, XW_K^{(i)}, XW_V^{(i)}) \nonumber , \\
    &= \text{softmax}\left(\frac{(XW_Q^{(i)})(XW_K^{(i)})^T}{\sqrt{d_k}}\right) (XW_V^{(i)}) , \\
    \label{eq:Q,K,V}
\end{aligned}
\end{equation}
\begin{equation}
\begin{aligned}
    \text{MHA}(X) &= \text{Concat}(\text{head}_1, \dots, \text{head}_h)W_O ,\\
    \label{eq: attention Q,K,V}
\end{aligned}
\end{equation}
\begin{equation}
    G = \sigma (XW_G), \\
    \label{eq: gated G}
\end{equation}
\begin{equation}
\begin{aligned}
    H &= \text{MHA}(X) \odot G ,\\
    \label{eq: output H}
\end{aligned}
\end{equation}

\textbf{ReZero Normalization}
ReZero incorporates a learnable parameter $\alpha$ into the residual connection, which is initialized to zero. The $l$-th layer is formulated as follows:
\begin{equation}
\begin{aligned}
    \hat{H}^{(l)} &= H^{(l-1)} + \alpha_{1}^{(l)} \cdot \text{GatedAttn}(H^{(l-1)}), \\
    \label{eq: rezero norm 1}
\end{aligned}
\end{equation}
\begin{equation}
\begin{aligned}
    H^{(l)} &= \hat{H}^{(l)} + \alpha_{2}^{(l)} \cdot \text{FFN}(\hat{H}^{(l)}), \\
    \label{eq: rezero norm 2}
\end{aligned}
\end{equation}

\subsection{Output Layer}

Following the $N$ Decoder Layers, the output node embeddings $H$ are passed through a linear projection layer to obtain Logits. Then,  we mask out invalid nodes (i.e., start or current nodes) to ensure validity in Equation~\ref{eq: mask logits}. 
The model computes the predicted probabilities for the nodes based on Equation~\ref{eq: output prob}.
\begin{equation}
    a_i = \begin{cases} 
    x_i & \forall i \in \{1 : l\} \text{ available nodes} \\ 
    -\infty & \text{otherwise}
    \end{cases}
    \label{eq: mask logits}
\end{equation}
\begin{equation}
    \mathbf{p} = \text{Softmax}(\mathbf{a})
    \label{eq: output prob}
\end{equation}

\subsection{Effect of Gated Attention}

Following~\citep{qiu2025gatedattentionlargelanguage}, we incorporate a gating mechanism into each attention layer, as formulated in Equation~\ref{eq: output H}. We conduct comparative experiments to investigate whether this gated attention improves the model's overall performance.
We evaluate a deep model ($D=42, W=128$) to isolate the impact of the gating mechanism. Holding all other hyperparameters constant, we compare models with and without gated attention. Our test set consists of TSP instances with node coordinates sampled from a uniform distribution. Specifically, it includes 10,000 instances for TSP100, and 128 instances for each TSP200/500/1000.

\begin{figure}[ht]
\centerline{\includegraphics[width=0.6\linewidth]{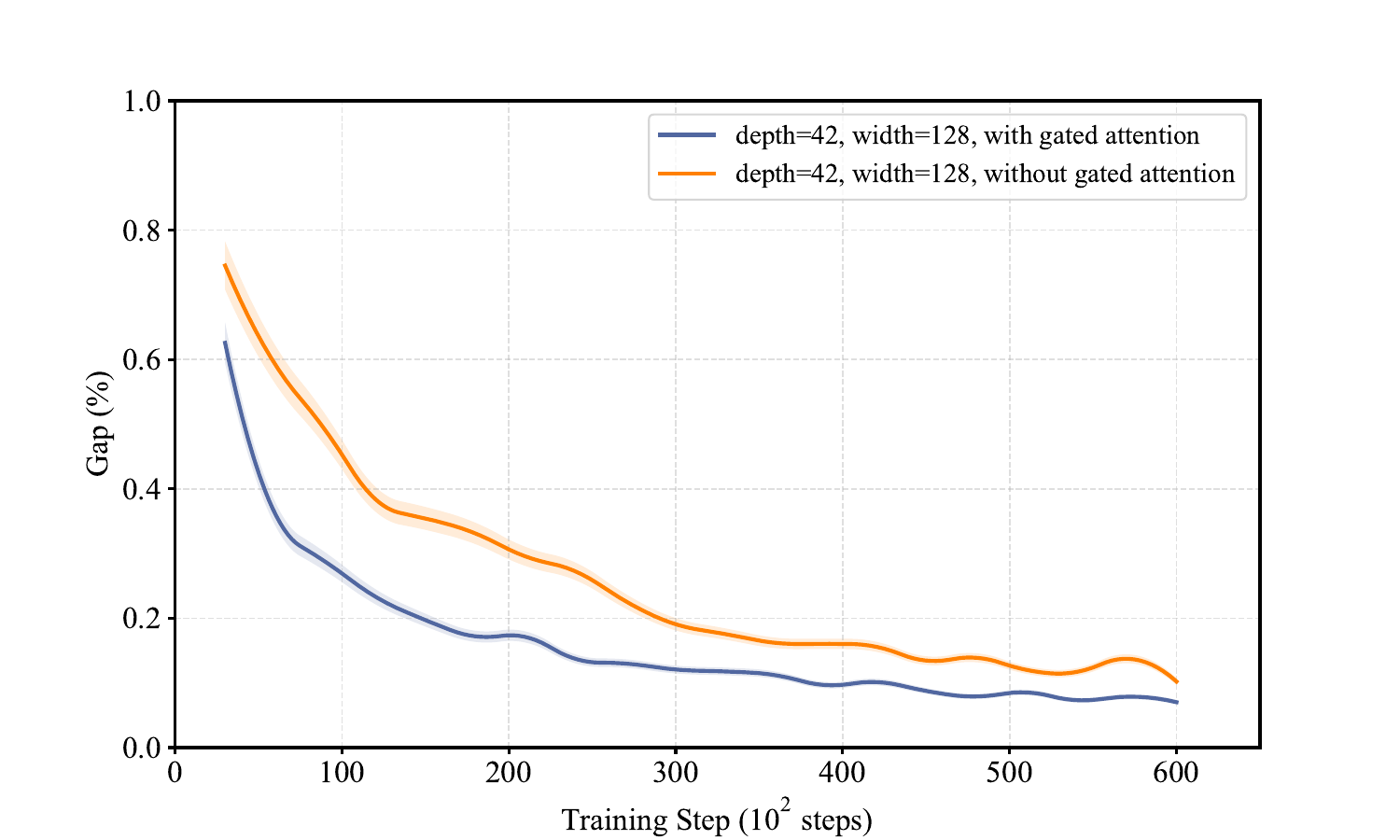}}
\caption{Convergence: w/ vs. w/o Gated Attention}
\label{fig: with and without gated}
\end{figure}

As illustrated in Figure~\ref{fig: with and without gated}, the model with Gated Attention ($D=42, W=128$) demonstrates faster convergence compared to its counterpart without it. Furthermore, it achieves superior performance both in-domain (0.070\% vs. 0.105\% gap) and in cross-scale generalization to larger instances ($N \ge 200$). Notably, on TSP1000, the optimality gap is reduced from 1.023\% to 0.869\%. These results indicate that Gated Attention not only accelerates training but also enhances overall solution quality.

\begin{table*}[ht]
  \caption{Performance Comparison: w/ Gated Attention and w/o Gated Attention.}
  \label{tb: with gated attention}
  \begin{center}
      \resizebox{0.7\columnwidth}{!}{
        \begin{tabular}{l|c|c|c|c }
          \toprule
            & TSP100 & TSP200 & TSP500 & TSP1000 \\ 
          \midrule
            Ours greedy (Depth=42,Width=128) w/ gated attention & 0.070\% & 0.178\% & 0.477\% & 0.869\% \\
            Ours greedy (Depth=42,Width=128) w/o gated attention & 0.105\% & 0.245\% & 0.614\% & 1.023\%\\
          \bottomrule
        \end{tabular}
        }
  \end{center}
\end{table*}

\subsection{Effect of ReZero Normalization}

To verify that ReZero Normalization~\citep{bachlechner2021rezero} stabilizes the training of deep models, we conduct an ablation study monitoring the loss magnitude during convergence as an indicator of training stability.
Specifically, using the same model configuration ($D=42, W=128$), we evaluate two variants: one with ReZero Normalization and one without. The model with ReZero Normalization utilizes an initial learning rate of $1.25 \times 10^{-4}$. In contrast, as the baseline (without ReZero) fails to converge at this learning rate, we employ a warmup strategy consisting of 50 steps with a start factor of 0.1 to ensure stability.

\begin{figure*}[ht]
\centerline{\includegraphics[width=0.6\linewidth]{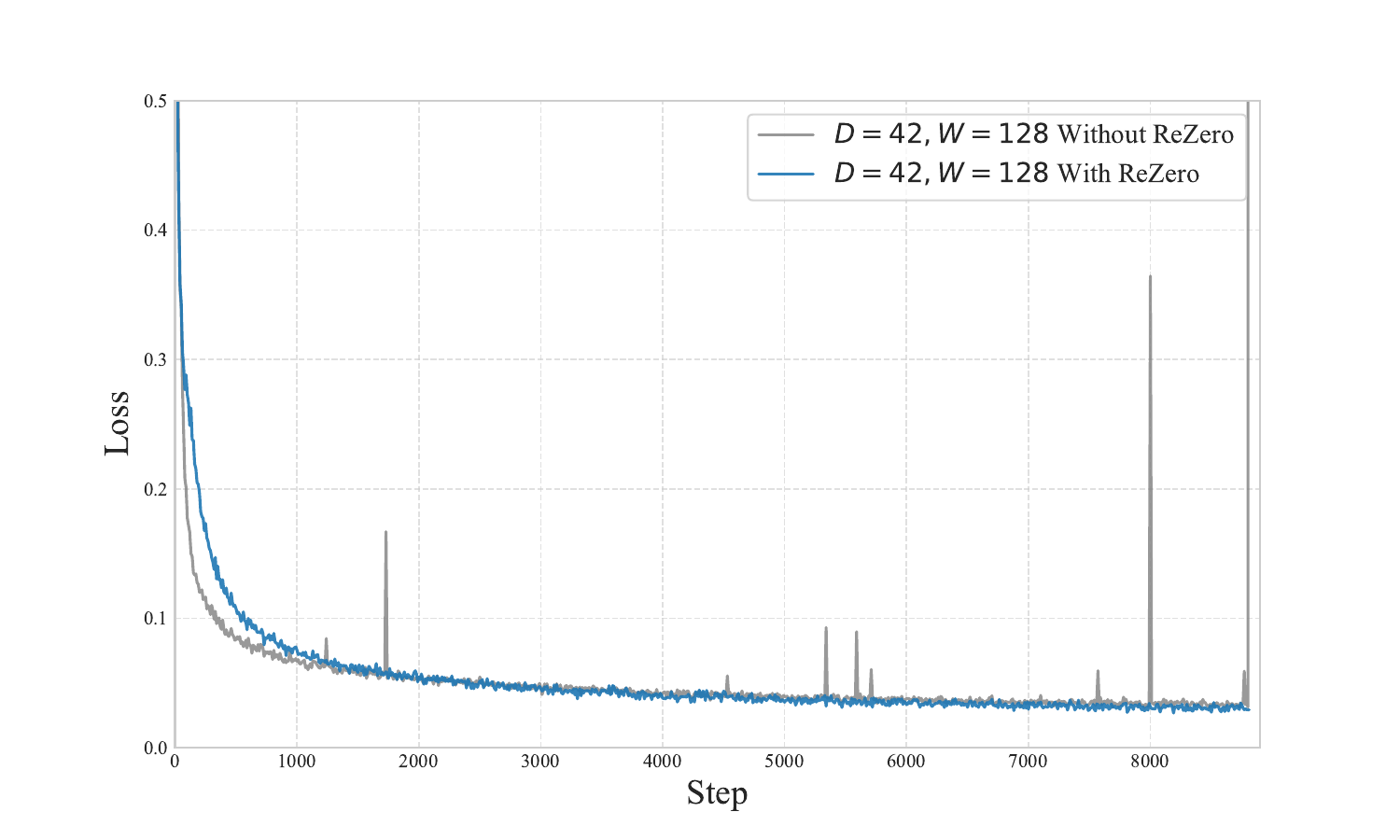}}
\caption{Convergence: w/ vs. w/o Rezero Normalization.}
\label{fig: rezero}
\end{figure*}

The training dynamics are visualized in Figure~\ref{fig: rezero}. From this figure, we can observe that the model without ReZero Normalization ($D=42, W=128$) exhibits frequent loss spikes and eventually diverges. In contrast, the model incorporating ReZero Normalization yields a smooth trajectory and stable convergence. These results confirm that ReZero Normalization is helpful in stabilizing the training of deep models.

\subsection{Effect of $\log(n)$}

Direct application to cross-scale instances often causes a shift in attention entropy, leading to performance degradation. To mitigate this, we follow~\citet{xiao2025improvinggeneralizationneuralvehicle} and add a $\log(n)$ correction to the attention weights during inference (Equation~\ref{eq: log n}). This adjustment aligns the attention distribution with that of the training phase, thereby restoring generalization capability. The formulation of the correction process is described as follows: 

\begin{equation}
    \text{Attn}(Q, K, V) = \sigma(Q K^T \cdot \log_{n_{tr}} n_{te}) V ,
\label{eq: log n}
\end{equation}
where $n_{tr} = 100$ represents the problem size of the training instances, and $n_{te}$ denotes the problem size of the testing instances.

To assess the impact of the $\log(n)$ term, we select three configurations: a baseline ($D=6, W=128$, 1.32M), a deep variant ($D=42, W=128$, 9.02M), and a wide variant ($D=6, W=512$, 21.00M). We evaluate these models on uniformly distributed TSP instances, using 10,000 samples for TSP100 and 128 samples for each larger scale (TSP200, TSP500, and TSP1000). The results are shown in Table~\ref{with and without log(n)}.

The results demonstrate that calibrating attention weights with a $\log(n)$ factor significantly enhances cross-scale generalization. On TSP1000 instances, this correction improved relative performance by 46.37\% for the baseline, 45.48\% for the deep variant, and 32.83\% for the wide variant.

\begin{table}[ht]
    \centering
    \caption{Performance Comparison: w/ and w/o $\log(n)$.}
    \label{with and without log(n)}

    \resizebox{0.6\linewidth}{!}{
    \begin{tabular}{llcccc}
        \toprule
        Setting & Model & TSP100 & TSP200 & TSP500 & TSP1000 \\
        \midrule
        \multirow{3}{*}{w/o $\log(n)$} 
        & Ours ($D=6,W=128$) & 0.464\% & 0.644\% & 1.645\% & 4.050\% \\
        & Ours ($D=6,W=256$)  & 0.190\% & 0.341\% & 0.880\% & 2.239\% \\
        & Ours ($D=42,W=128$) & 0.070\% & 0.184\% & 0.532\% & 1.594\% \\
        \midrule
        \multirow{3}{*}{w/ $\log(n)$}    
        & Ours ($D=6,W=128$)  & 0.464\% & 0.669\% & 1.559\% & 2.172\% \\
        & Ours ($D=6,W=256$)  & 0.264\% & 0.498\% & 0.969\% & 1.504\% \\
        & Ours ($D=42,W=128$) & 0.070\% & 0.178\% & 0.477\% & 0.869\% \\
        \bottomrule
    \end{tabular}
    }
\end{table}

\newpage

\clearpage

\newpage

\section{Training Hyperparameters, Hardware, and Data}
\label{apl: training details}

\subsection{Training Hyperparameters and Hardware}

The detailed hyperparameters are provided in Table~\ref{tb: training parameters}. We maintain consistent configurations across all 12 models to ensure a fair comparison.
\begin{table*}[ht]
  \caption{Training hyperparameters.}
  \label{tb: training parameters}
  \begin{center}
      \resizebox{0.3\columnwidth}{!}{
        \begin{tabular}{l|c}
          \toprule
           parameters & values \\ 
          \midrule
           Batch size & 1024 \\
           Total training steps & 60,000 \\
           optimizer & AdamW \\
           Weight decay & 0.01 \\
           Learning rate & $1.25 \times 10^{-4}$ \\
           Scheduler & Exponential \\
           Decay gamma & 0.997 \\
           Decay step & 100 \\   
          \bottomrule
        \end{tabular}
        }
  \end{center}
\end{table*}

All experiments are conducted on NVIDIA RTX 4090 GPUs equipped with 24GB of memory. Detailed hardware assignments for each specific model are provided in ~ Table\ref{tb: device detail}.

\begin{table}[ht]
    \caption{Hardware Specifications.}
    \label{tb: device detail}
    \centering
    \resizebox{0.5\linewidth}{!}{ 
        \begin{tabular}{cc @{\hspace{1.5cm}} cc}
            \toprule
            Configuration & Device & Configuration & Device \\ 
            \midrule
            $D=6,W=128$  & 4090$\times 2$ & $D=24,W=128$ & 4090$\times 2$ \\
            $D=6,W=256$  & 4090$\times 2$ & $D=24,W=256$ & 4090$\times 8$ \\
            $D=6,W=512$  & 4090$\times 8$ & $D=24,W=512$ & 4090$\times 8$ \\
            \midrule
            $D=12,W=128$ & 4090$\times 2$ & $D=42,W=128$ & 4090$\times 4$ \\
            $D=12,W=256$ & 4090$\times 2$ & $D=42,W=256$ & 4090$\times 8$ \\
            $D=12,W=512$ & 4090$\times 4$ & $D=42,W=512$ & 4090$\times 8$ \\
            \bottomrule
        \end{tabular}
    }
\end{table}

\subsection{Training Data}

\label{apl: training data}

To accurately assess the model's intrinsic capabilities and eliminate performance degradation caused by overfitting to limited datasets, it is essential to treat the training set as sufficiently large. Therefore, we generate a comprehensive training dataset comprising 60 million TSP100 instances, with node coordinates sampled from a uniform distribution in the interval $[0,1]^2$. The ground truth label for each instance is obtained using the Concorde solver~\citep{concorde}. Instead of conventional epoch-based training where the model iterates over the dataset multiple times, we adopt a step-based training strategy. In this approach, the model processes each instance exactly once. We train the models for a total of 60,000 steps (defining each batch update as a single step), at which point we observe that all 12 models have converged.

\begin{figure}[ht]
\centerline{\includegraphics[width=0.7\linewidth]{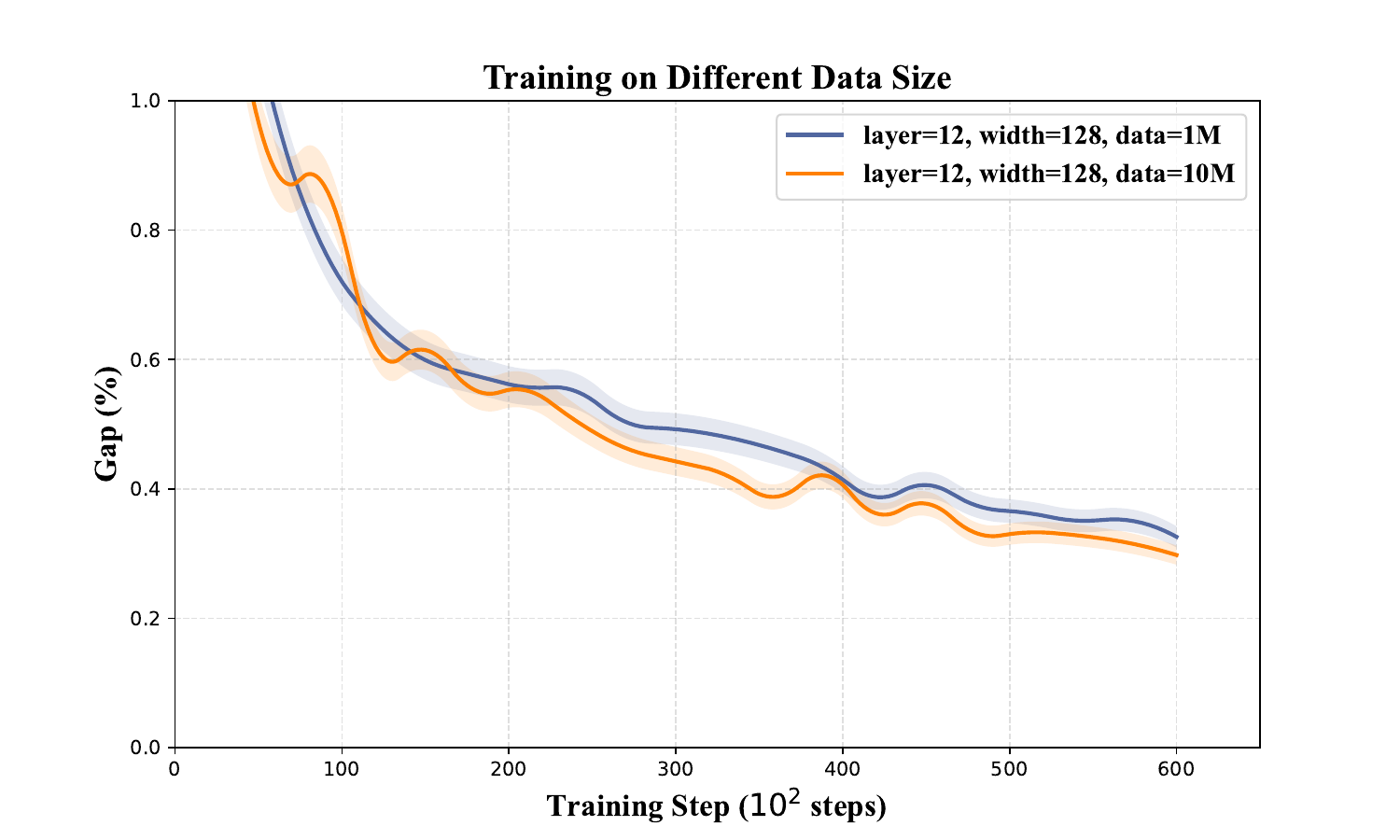}}
\caption{Convergence comparison of the model ($\text{Depth}=12, \text{Width}=128$) trained on datasets of varying sizes (1M vs. 10M). The model trained on 10 million training samples (orange curve) demonstrates faster convergence and achieves a lower optimality gap compared to the 1 million sample baseline.}
\label{fig: train data size}
\end{figure}

To validate the effectiveness of this large-scale data approach, we conduct an ablation study to analyze the impact of training data volume. We compare the performance of a fixed model architecture ($\text{Depth}=12, \text{Width}=128$) trained on a subset of 1 million samples against one trained on 10 million samples. The results are shown in Figure~\ref{fig: train data size} and Table~\ref{tb: different training data size}. 

As evidenced by the convergence curves in Figure~\ref{fig: train data size}, scaling the training dataset from 1M to 10M effectively accelerates the convergence process. Furthermore, as detailed in Table~\ref{tb: different training data size}, this increase in data volume yields superior optimality gaps on TSP100. Crucially, the larger dataset significantly enhances cross-scale generalization capabilities when testing on larger problem sizes (TSP200, TSP500, and TSP1000). These results demonstrate that a sufficiently large, non-repeating training set is critical for maximizing performance and generalization, justifying our adoption of the massive 60-million instance dataset for the final training process.

\begin{table*}[ht]
  \caption{Performance comparison on TSP instances of varying sizes (TSP100 to TSP1000) using models trained on datasets with sizes of 1M and 10M.}
  \label{tb: different training data size}
  \begin{center}
      \resizebox{0.8\columnwidth}{!}{
        \begin{tabular}{l|c|c|c|c|c }
          \toprule
             & Training Dataset Size  & TSP100 & TSP200 & TSP500 & TSP1000 \\ 
          \midrule
            Ours greedy (Depth=12,Width=128) & 1M  & 0.337\% & 0.615\% & 1.062\% & 1.606\% \\
            Ours greedy (Depth=12,Width=128) & 10M  & 0.302\% & 0.461\% & 0.972\% & 1.357\% \\
          \bottomrule
        \end{tabular}
        }
  \end{center}
\end{table*}

\clearpage

\newpage

\section{Results using Greedy Search}
\label{Performance using Greedy Search}

\subsection{In-domain Performance}
\label{In-domain performance}

Table~\ref{tb: tsp100-result-indomain-TSP100} presents the evaluation results of various model configurations on 10,000 in-domain TSP100 Uniform instances, which are used to fit the power law curves in Figure~\ref{fig: 7 curves}.

\begin{table}[ht]
\centering
\caption{Performance of different configurations on TSP 100 instances.}
\label{tb: tsp100-result-indomain-TSP100}
\resizebox{0.48\textwidth}{!}{%
\begin{tabular}{l|l|c c}
\toprule[0.5mm]
 & & \multicolumn{2}{c}{TSP 100} \\
 & & Gap & Time \\
 \cmidrule{2-4}
 \multirow{13}{*}{Uniform} & LKH3 &  0.000\% & 56m \\
 \cmidrule{2-4}
 & Ours greedy (Depth=6,Width=128) & 0.464\% & 0.34m \\
 & Ours greedy (Depth=6,Width=256) & 0.264\% & 0.75m \\
 & Ours greedy (Depth=6,Width=512) & 0.190\% & 1.7m \\
 & Ours greedy (Depth=12,Width=128) & 0.225\% & 0.62m \\
 & Ours greedy (Depth=12,Width=256) & 0.115\% & 1.4m \\
 & Ours greedy (Depth=12,Width=512) & 0.083\% & 3.3m \\
 & Ours greedy (Depth=24,Width=128) & 0.106\% & 1.2m \\
 & Ours greedy (Depth=24,Width=256) & 0.075\% & 2.8m \\
 & Ours greedy (Depth=24,Width=512) & 0.052\% & 6.4m \\
 & Ours greedy (Depth=42,Width=128) & 0.070\% & 2.0m \\
 & Ours greedy (Depth=42,Width=256) & 0.052\% & 4.8m \\
 & Ours greedy (Depth=42,Width=512) & 0.035\% & 11m \\
\bottomrule[0.5mm]
\end{tabular}%
}
\end{table}

\clearpage

\newpage

\subsection{Out-of-domain Performance}

Tables~\ref{tb: appendix-result-Explosion},~\ref{tb: appendix-result-Implosion}, and~\ref{tb: appendix-result-cluster} show the performance of different model configurations across 15 out-of-domain test sets, spanning four problem sizes $N \in \{100,200,500,1000\}$ and four distributions $\text{Distribution} \in \{\text{Uniform, Explosion, Implosion, and Cluster}\}$ similar to~\citep{fang2024invit}. Note that all models are trained exclusively on TSP100-Uniform instances to evaluate their cross-distribution and cross-size generalization capabilities. The overall trends can be summarized as follows:

\textbf{Impact of Model Capacity} There is a generally consistent correlation between model size and solution quality across all distributions. Increasing both the depth (from 6 to 42) and width (from 128 to 512) of the model generally reduces the optimality gap. The largest configuration (Depth=42, Width=512) achieves the lowest gaps across all test sets. For example, in the Uniform distribution (TSP1000), the gap drops from 2.172\% (smallest model) to 0.576\% (largest model). This performance gain comes with a computational trade-off. The inference time scales significantly with model size, increasing from seconds to nearly half an hour for the largest instances.

\textbf{Generalization Across Problem Sizes} The models demonstrate the ability to generalize to problem sizes larger than those seen during training (training on $N=100$, testing up to $N=1000$). Across all distributions, the optimality gap widens as the problem size increases from 100 to 1000. Despite this degradation, the larger models maintain relatively low gaps ($<$1.3\%) even on TSP1000 for Uniform, Explosion, and Implosion distributions, indicating strong zero-shot generalization capabilities for size.

\textbf{Robustness Across Data Distributions} The models show varying degrees of robustness to distributional shifts. The performance on Explosion (Table~\ref{tb: appendix-result-Explosion}) and Implosion (Table~\ref{tb: appendix-result-Implosion}) is comparable to the in-domain Uniform distribution (Table~\ref{tb: tsp-large-result}). This suggests the model captures structural patterns common to these distributions. The Cluster distribution (Table~\ref{tb: appendix-result-cluster}) proves to be the most challenging out-of-domain scenario.

\begin{table*}[htbp]
\centering
\caption{Performance of different configurations on TSP 200, 500, and 1000 instances.}
\label{tb: tsp-large-result}
\resizebox{0.7\textwidth}{!}{%
\begin{tabular}{l|l|c c|c c|c c}
\toprule[0.5mm]
 & & \multicolumn{2}{c|}{TSP 200} & \multicolumn{2}{c|}{TSP 500} & \multicolumn{2}{c}{TSP 1000} \\
 & & Gap & Time & Gap & Time & Gap & Time \\
 \cmidrule{2-8}
 \multirow{13}{*}{Uniform} & LKH3 & 0.000\% & 17m & 0.000\% & 50m & 0.000\% & 10h  \\
 \cmidrule{2-8}
 & Ours greedy (Depth=6,Width=128) & 0.669\% & 0.035m & 1.559\% & 0.23m & 2.172\% & 1.6m  \\
 & Ours greedy (Depth=6,Width=256) & 0.498\% & 0.059m & 0.969\% & 0.48m & 1.504\% & 3.2m  \\
 & Ours greedy (Depth=6,Width=512) & 0.375\% & 0.1m & 0.856\% & 0.75m & 1.291\% & 4.2m  \\
 & Ours greedy (Depth=12,Width=128) & 0.550\% & 0.054m & 0.803\% & 0.44m & 1.357\% & 3.1m  \\
 & Ours greedy (Depth=12,Width=256) & 0.258\% & 0.1m & 0.572\% & 0.93m & 1.025\% & 6.2m  \\
 & Ours greedy (Depth=12,Width=512) & 0.216\% & 0.18m & 0.490\% & 1.5m & 0.856\% & 8.2m  \\
 & Ours greedy (Depth=24,Width=128) & 0.262\% & 0.092m & 0.559\% & 0.85m & 0.976\% & 6.1m  \\
 & Ours greedy (Depth=24,Width=256) & 0.191\% & 0.18m & 0.469\% & 1.8m & 0.765\% & 12m  \\
 & Ours greedy (Depth=24,Width=512) & 0.130\% & 0.035m & 0.453\% & 2.9m & 0.841\% & 16m  \\
 & Ours greedy (Depth=42,Width=128) & 0.178\% & 0.15m & 0.477\% & 1.5m & 0.869\% & 11m  \\
 & Ours greedy (Depth=42,Width=256) & 0.121\% & 0.31m & 0.433\% & 3.2m & 0.739\% & 21m  \\
 & Ours greedy (Depth=42,Width=512) & 0.110\% & 0.61m & 0.293\% & 5.1m & 0.576\% & 28m  \\
\bottomrule[0.5mm]
\end{tabular}%
}
\end{table*}

\begin{table*}[htbp]
\centering
\caption{Performance of different configurations on TSP instances with different problem sizes and the distribution of explosion. Ground truth solutions are obtained via LKH3 for the test dataset.}
\label{tb: appendix-result-Explosion}
\resizebox{0.85\textwidth}{!}{%
\begin{tabular}{l|l|c c|c c|c c|c c}
\toprule[0.5mm]
 & & \multicolumn{2}{c|}{TSP 100} & \multicolumn{2}{c|}{TSP 200} & \multicolumn{2}{c|}{TSP 500} & \multicolumn{2}{c}{TSP 1000} \\
 & & Gap & Time & Gap & Time & Gap & Time & Gap & Time \\
 \cmidrule{2-10}
 \multirow{13}{*}{Explosion} & LKH3 &  0.000\% & 56m & 0.000\% & 17m & 0.000\% & 50m & 0.000\% & 10h \\
 \cmidrule{2-10}
 & Ours greedy (Depth=6,Width=128) & 0.496\% & 0.35m & 0.767\% & 0.037m & 1.905\% & 0.23m & 3.819\% & 1.6m  \\
 & Ours greedy (Depth=6,Width=256) & 0.301\% & 0.75m & 0.506\% & 0.059m & 1.143\% & 0.48m & 2.430\% & 3.2m  \\
 & Ours greedy (Depth=6,Width=512) & 0.222\% & 1.7m & 0.403\% & 0.1m & 0.974\% & 0.75m & 2.029\% & 4.2m  \\
 & Ours greedy (Depth=12,Width=128) & 0.245\% & 0.64m & 0.420\% & 0.055m & 1.044\% & 0.44m & 1.987\% & 3.1m  \\
 & Ours greedy (Depth=12,Width=256) & 0.123\% & 1.4m & 0.239\% & 0.1m & 0.782\% & 0.93m & 1.399\% & 6.2m  \\
 & Ours greedy (Depth=12,Width=512) & 0.093\% & 3.3m & 0.172\% & 0.19m & 0.626\% & 1.5m & 1.223\% & 8.3m  \\
 & Ours greedy (Depth=24,Width=128) & 0.107\% & 1.2m & 0.167\% & 0.096m & 0.610\% & 0.86m & 1.249\% & 6.1m  \\
 & Ours greedy (Depth=24,Width=256) & 0.088\% & 2.8m & 0.164\% & 0.19m & 0.547\% & 1.8m & 1.010\% & 12m  \\
 & Ours greedy (Depth=24,Width=512) & 0.086\% & 6.5m & 0.121\% & 0.35m & 0.529\% & 2.9m & 1.132\% & 16m  \\
 & Ours greedy (Depth=42,Width=128) & 0.144\% & 2.1m & 0.154\% & 0.15m & 0.559\% & 1.5m & 1.271\% & 11m  \\
 & Ours greedy (Depth=42,Width=256) & 0.056\% & 4.8m & 0.114\% & 0.31m & 0.467\% & 3.2m & 0.958\% & 21m  \\
 & Ours greedy (Depth=42,Width=512) & 0.039\% & 11m & 0.081\% & 0.6m & 0.351\% & 5.1m & 0.670\% & 28m  \\

\bottomrule[0.5mm]
\end{tabular}%
}
\end{table*}

\begin{table*}[htbp]
\centering
\caption{Performance of different configurations on TSP instances with different problem sizes and the distribution of implosion. Ground truth solutions are obtained via LKH3 for the test dataset.}
\label{tb: appendix-result-Implosion}
\resizebox{0.85\textwidth}{!}{%
\begin{tabular}{l|l|c c|c c|c c|c c}
\toprule[0.5mm]
 & & \multicolumn{2}{c|}{TSP 100} & \multicolumn{2}{c|}{TSP 200} & \multicolumn{2}{c|}{TSP 500} & \multicolumn{2}{c}{TSP 1000} \\
 & & Gap & Time & Gap & Time & Gap & Time & Gap & Time \\
 \cmidrule{2-10}
 \multirow{13}{*}{Implosion} & LKH3 &  0.000\% & 56m & 0.000\% & 17m & 0.000\% & 50m & 0.000\% & 10h  \\
 \cmidrule{2-10}
 & Ours greedy (Depth=6,Width=128) & 0.895\% & 0.35m & 1.424\% & 0.035m & 2.909\% & 0.23m & 3.507\% & 1.7m  \\
 & Ours greedy (Depth=6,Width=256) & 0.529\% & 0.74m & 0.646\% & 0.06m & 1.660\% & 0.48m & 2.580\% & 3.2m  \\
 & Ours greedy (Depth=6,Width=512) & 0.436\% & 1.7m & 0.634\% & 0.1m & 1.494\% & 0.75m & 2.164\% & 4.2m  \\
 & Ours greedy (Depth=12,Width=128) & 0.491\% & 0.63m & 0.824\% & 0.055m & 1.725\% & 0.44m & 2.246\% & 3.1m  \\
 & Ours greedy (Depth=12,Width=256) & 0.307\% & 1.4m & 0.477\% & 0.1m & 1.299\% & 0.93m & 1.798\% & 6.2m  \\
 & Ours greedy (Depth=12,Width=512) & 0.229\% & 3.3m & 0.399\% & 0.18m & 1.171\% & 1.5m & 1.482\% & 8.3m  \\
 & Ours greedy (Depth=24,Width=128) & 0.260\% & 1.2m & 0.351\% & 0.092m & 0.937\% & 0.86m & 1.580\% & 6.1m  \\
 & Ours greedy (Depth=24,Width=256) & 0.236\% & 2.8m & 0.449\% & 0.19m & 1.278\% & 1.8m & 1.419\% & 12m  \\
 & Ours greedy (Depth=24,Width=512) & 0.199\% & 6.5m & 0.296\% & 0.36m & 0.918\% & 2.9m & 1.360\% & 16m  \\
 & Ours greedy (Depth=42,Width=128) & 0.270\% & 2.1m & 0.352\% & 0.15m & 1.127\% & 1.5m & 1.685\% & 11m  \\
 & Ours greedy (Depth=42,Width=256) & 0.168\% & 4.8m & 0.282\% & 0.31m & 0.867\% & 3.2m & 1.261\% & 21m  \\
 & Ours greedy (Depth=42,Width=512) & 0.103\% & 11m & 0.284\% & 0.6m & 0.745\% & 5.1m & 1.023\% & 28m  \\
\bottomrule[0.5mm]
\end{tabular}%
}
\end{table*}

\begin{table*}[htbp]
\centering
\caption{Performance of different configurations on TSP instances with different problem sizes and the distribution of clusters. Ground truth solutions are obtained via LKH3 for the test dataset.}
\label{tb: appendix-result-cluster}
\resizebox{0.85\textwidth}{!}{%
\begin{tabular}{l|l|c c|c c|c c|c c}
\toprule[0.5mm]
 & & \multicolumn{2}{c|}{TSP 100} & \multicolumn{2}{c|}{TSP 200} & \multicolumn{2}{c|}{TSP 500} & \multicolumn{2}{c}{TSP 1000} \\
 & & Gap & Time & Gap & Time & Gap & Time & Gap & Time \\
 \cmidrule{2-10}
 \multirow{13}{*}{Cluster} & LKH3 &  0.000\% & 56m & 0.000\% & 17m & 0.000\% & 50m & 0.000\% & 10h  \\
 \cmidrule{2-10}
 & Ours greedy (Depth=6,Width=128) & 2.742\% & 0.34m & 2.440\% & 0.035m & 4.018\% & 0.23m & 6.639\% & 1.6m  \\
 & Ours greedy (Depth=6,Width=256) & 1.253\% & 0.74m & 1.240\% & 0.063m & 2.345\% & 0.48m & 4.276\% & 3.2m  \\
 & Ours greedy (Depth=6,Width=512) & 1.377\% & 1.7m & 1.164\% & 0.1m & 2.002\% & 0.75m & 3.542\% & 4.2m  \\
 & Ours greedy (Depth=12,Width=128) & 1.134\% & 0.63m & 1.178\% & 0.055m & 1.943\% & 0.44m & 3.539\% & 3.1m  \\
 & Ours greedy (Depth=12,Width=256) & 1.136\% & 1.4m & 1.111\% & 0.1m & 1.603\% & 0.93m & 2.543\% & 6.2m  \\
 & Ours greedy (Depth=12,Width=512) & 0.538\% & 3.3m & 0.596\% & 0.18m & 1.121\% & 1.5m & 2.139\% & 8.3m  \\
 & Ours greedy (Depth=24,Width=128) & 0.584\% & 1.2m & 0.634\% & 0.092m & 1.286\% & 0.85m & 2.185\% & 6.1m  \\
 & Ours greedy (Depth=24,Width=256) & 0.853\% & 2.8m & 0.854\% & 0.19m & 1.193\% & 1.8m & 1.691\% & 12m  \\
 & Ours greedy (Depth=24,Width=512) & 0.424\% & 6.5m & 0.438\% & 0.35m & 0.775\% & 2.9m & 1.873\% & 16m  \\
 & Ours greedy (Depth=42,Width=128) & 1.126\% & 2.1m & 0.775\% & 0.15m & 1.322\% & 1.5m & 2.489\% & 11m  \\
 & Ours greedy (Depth=42,Width=256) & 0.437\% & 4.8m & 0.628\% & 0.32m & 0.729\% & 3.2m & 1.561\% & 21m  \\
 & Ours greedy (Depth=42,Width=512) & 0.316\% & 11m & 0.383\% & 0.6m & 0.694\% & 5.1m & 1.253\% & 28m  \\
 
\bottomrule[0.5mm]
\end{tabular}%
}
\end{table*}

\newpage

\section{Compute Budget and Inference Time}
\label{Compute Budget and Inference Time}

While the theoretical compute budget (e.g., FLOPs) is determined by the model architecture, the real inference time is heavily influenced by hardware specifications and batch sizes. In this section, we investigate the relationship between the theoretical compute budget and the actual wall-clock inference time.
We evaluate two distinct model configurations: a shallow-wide variant ($D=12, W=512$, 41.46M) and a deep-narrow counterpart ($D=24, W=256$, 20.62M). We measure the total inference latency across varying batch sizes. All experiments are conducted on a single NVIDIA RTX 4090 GPU with 24GB of memory.

The results shown in Table~\ref{tb: time with different bs} reveal that at small batch sizes (e.g., 50), the GPU remains under-utilized. Despite requiring less compute budget per solution (434.4 vs. 843.2 GFLOPs), the deep-narrow model ($D=24, W=256$) is bottlenecked by the sequential dependency of its layers, leading to higher latency due to limited parallelism. However, as the batch size increases (e.g.， $>100$) and the GPU reaches saturation, the theoretical efficiency of the deep-narrow model is realized, outperforming the shallow-wide baseline ($D=12, W=512$). Consequently, we conclude that inference latency is highly hardware-dependent and exhibits a non-linear correlation with the theoretical compute budget.

\begin{table*}[ht]
  \caption{Inference time with different batch sizes.}
  \label{tb: time with different bs}
  \begin{center}
      \resizebox{0.8\columnwidth}{!}{
        \begin{tabular}{l|c|c|c|c }
          \toprule
             & Compute per solution & Batch size =50 & Batch size =100 & Batch size =1000 \\ 
          \midrule
            Ours greedy (Depth=12,Width=512) & 843.2 GFLOPs & 4.0m & 3.4m & 3.3m \\
            Ours greedy (Depth=24,Width=256) & 434.4 GFLOPs &  4.4m & 3.2m & 2.8m \\
          \bottomrule
        \end{tabular}
        }
  \end{center}
\end{table*}

\newpage

\section{Results using Beam Search}
\label{app: Inference using Beam Search}

To trade inference-time compute for enhanced solution quality, we employ the inference method beam search. During inference, we modulate the search space by varying the beam size from 1 to 64, and evaluate the performance of the trained 12 models on 10,000 TSP100 instances. The results are shown in Table~\ref{tb: beam search}.

From these results, we can observe a strict monotonic improvement in solution quality across all models as the compute budget increases. Notably, the shallow model ($D=6, W=128$) requires a beam size of 8 to surpass the greedy performance of the deeper model ($D=42, W=128$), suggesting that increased model depth effectively compensates for the need for extensive search during inference. Furthermore, depth yields a higher performance ceiling than width; the deep configuration ($D=42, W=128$) outperforms the wider variant ($D=6, W=512$) under the same beam size $=$ 64 ($0.0001\%$ vs. $0.0007\%$). Remarkably, the largest model ($D=42, W=512$) achieves a near-optimal gap of less than $10^{-6} $\%  with a beam size of 64.

\begin{table}[h]
    \centering
    \caption{The inference performance of different model configurations with different beam sizes on the TSP100 (10,000 instances). When Beam = 1, beam search degenerates into greedy search.}
    \resizebox{0.8\textwidth}{!}{%
        \begin{tabular}{lrrrrrrr}
            \toprule
             & greedy & {beam=2} & {beam=4} & {beam=8} & {beam=16} & {beam=32} & {beam=64} \\
            \midrule
            Ours ($D=6,W=128$)  & 0.4640\% & 0.2456\% & 0.1155\% & 0.0571\% & 0.0269\% & 0.0127\% & 0.0059\% \\
            Ours ($D=6,W=256$)  & 0.2635\% & 0.1397\% & 0.0647\% & 0.0289\% & 0.0125\% & 0.0008\% & 0.0017\% \\
            Ours ($D=6,W=512$)  & 0.1895\% & 0.0904\% & 0.0359\% & 0.0138\% & 0.0050\% & 0.0019\% & 0.0007\% \\
            \midrule
            Ours ($D=12,W=128$) & 0.2253\% & 0.1122\% & 0.0512\% & 0.0223\% & 0.0088\% & 0.0032\% & 0.0010\% \\
            Ours ($D=12,W=256$) & 0.1149\% & 0.0523\% & 0.0200\% & 0.0069\% & 0.0024\% & 0.0008\% & 0.0002\% \\
            Ours ($D=12,W=512$) & 0.0827\% & 0.0354\% & 0.0133\% & 0.0044\% & 0.0016\% & 0.0005\% & 0.0002\% \\
            \midrule
            Ours ($D=24,W=128$) & 0.1064\% & 0.0489\% & 0.0177\% & 0.0062\% & 0.0021\% & 0.0007\% & 0.0002\% \\
            Ours ($D=24,W=256$) & 0.0746\% & 0.0312\% & 0.0117\% & 0.0038\% & 0.0011\% & 0.0003\% & 0.0001\% \\
            Ours ($D=24,W=512$) & 0.0520\% & 0.0211\% & 0.0069\% & 0.0017\% & 0.0004\% & 0.0001\% & 0.0000\% \\
            \midrule
            Ours ($D=42,W=128$) & 0.0703\% & 0.0313\% & 0.0104\% & 0.0035\% & 0.0010\% & 0.0002\% & 0.0001\% \\
            Ours ($D=42,W=256$) & 0.0524\% & 0.0214\% & 0.0007\% & 0.0021\% & 0.0006\% & 0.0001\% & 0.0000\% \\
            Ours ($D=42,W=512$) & 0.0353\% & 0.0127\% & 0.0007\% & 0.0008\% & 0.0002\% & 0.0000\% & 0.0000\% \\
            \bottomrule
        \end{tabular}%
    }
    \label{tb: beam search}
\end{table}

\newpage

\section{Results using RRC}
\label{app: Inference using RRC}

In addition to beam search, we also try the inference method Random reconstruction (RRC)~\citep{luo2023neural} to trade off the compute budget for better performance. RRC uses the model to randomly destroy and reconstruct partial solutions.
We focus our analysis on four representative configurations: a baseline ($D=6, W=128$), its deeper ($D=42, W=128$) and wider ($D=6, W=512$) counterparts, and the largest model ($D=42, W=512$). We evaluate these models' in-domain performance on 1,000 TSP100 instances, and cross-scale generalization ability on 16 TSP1000 instances. The results are presented in Table~\ref{tab:tsp_results}.

The results show that increasing the compute budget during inference significantly enhances the model's out-of-distribution generalization. On TSP1000, the baseline model ($D=6, W=128$) reduces its optimality gap from 2.2667\% (Greedy) to 0.4785\% (RRC1000), representing a 4.73$\times$ performance gain. Notably, this benefit scales with model capacity: the shallow-wide ($D=6, W=512$) and deep-narrow ($D=48, W=128$) variants achieve relative improvements of 6.04$\times$ and 6.15$\times$, respectively. Furthermore, by leveraging the RRC strategy, our largest model achieves an optimality gap of less than 0.1\% on TSP1000. These results demonstrate the substantial efficacy of test-time scaling in improving cross-scale generalization.

\begin{table}[ht]
  \centering
  \caption{Results on TSP100 and TSP1000 with different budgets.}
  \resizebox{0.7\textwidth}{!}{%
  \label{tab:tsp_results}
  
  \begin{tabular}{lrrrrrr}
    \toprule
    TSP100 (1,000 instances) & {Greedy} & {RRC 100} & {RRC 200} & {RRC 500} & {RRC 800} & {RRC 1000} \\
    \midrule
    Ours ($D=6,W=128$)  & 0.4228\% & 0.0103\% & 0.0068\% & 0.0040\% & 0.0030\% & 0.0024\% \\
    Ours ($D=6,W=512$)  & 0.1824\% & 0.0033\% & 0.0018\% & 0.0007\% & 0.0006\% & 0.0005\% \\
    Ours ($D=42,W=128$) & 0.0656\% & 0.0015\% & 0.0012\% & 0.0004\% & 0.0004\% & 0.0004\% \\
    Ours ($D=42,W=512$) & 0.0332\% & 0.0008\% & 0.0006\% & 0.0003\% & 0.0002\% & 0.0002\% \\
    \bottomrule
  \end{tabular}
    }

\resizebox{0.7\textwidth}{!}{%
  \begin{tabular}{lrrrrrr}
    \toprule
    {TSP1000 (16 instances)} & {Greedy} & {RRC 100} & {RRC 200} & {RRC 500} & {RRC 800} & {RRC 1000} \\
    \midrule
    Ours ($D=6,W=128$)  & 2.2667\% & 0.7914\% & 0.6677\% & 0.5207\% & 0.4907\% & 0.4785\% \\
    Ours ($D=6,W=512$)  & 1.5174\% & 0.4498\% & 0.3886\% & 0.2861\% & 0.2651\% & 0.2511\% \\
    Ours ($D=42,W=128$) & 0.9691\% & 0.2771\% & 0.2328\% & 0.1703\% & 0.1590\% & 0.1575\% \\
    Ours ($D=42,W=512$) & 0.5014\% & 0.1258\% & 0.0930\% & 0.0852\% & 0.0765\% & 0.0755\% \\
    \bottomrule
  \end{tabular}
  }
\end{table}

\newpage

\section{Additional Analysis: Scaling Depth vs. Width}
\label{Additional Analysis: Scaling Depth vs. Width}

\subsection{Studies on Node Representations}

To investigate the impact of different scaling strategies on node representations, we extract the node embeddings from the final decoder layer of each model, and examine the model's separability to distinguish the next optimal node. We employ two methods for analysis: (1) \textbf{Principal Component Analysis (PCA)} to visualize the feature distribution in the principal subspace, (2) \textbf{Feature Map Analysis}, which provides an intuitive measure of the model's spatial compression regarding non-optimal (interfering) nodes based on pairwise cosine similarity.

We use the baseline model ($D=6, W=128$) and compare it against two variants: a deep architecture ($D=42, W=128$) and a wide architecture ($D=6, W=256$). At each decoding step, given the fixed start and current nodes, we evaluate the model's ability to identify the optimal next node from the unvisited set. To facilitate visualization, the indices of unvisited nodes are reordered according to the ground truth solution, such that the optimal node is always positioned at the first index.

\subsection{PCA Analysis}

\begin{figure}[ht]
    \centering   \centerline{\includegraphics[width=1.0\linewidth]{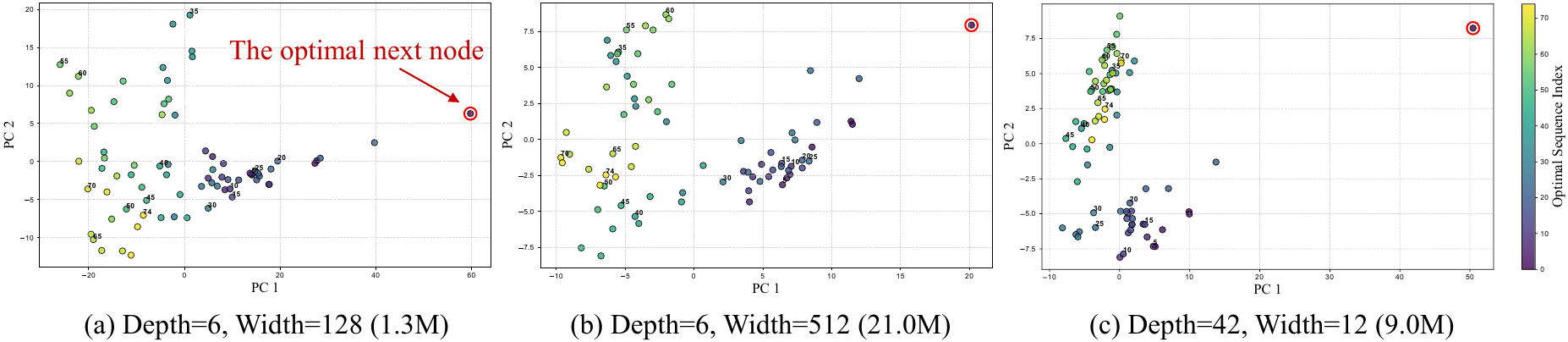}}
        \caption{\textbf{2D-PCA visualization for features of available nodes} \textbf{(a):} Background nodes exhibit a dispersed distribution with multiple isolated outliers. \textbf{(b):} The background distribution remains scattered. \textbf{(c):} Background nodes form a compact cluster, with the optimal node clearly separated from the background.}
        \label{fig: PCA}
\end{figure}

To evaluate the model's ability to distinguish the optimal node from other nodes, we project the node embeddings extracted from the decoder's final attention layer onto a 2D plane via Principal Component Analysis (PCA), and we visualize the distinct topological structures learned by different configurations (Figure~\ref{fig: PCA}). The comparison reveals a fundamental difference in how width and depth shape the feature space. While the wider model ($D=6, W=512$) separates the optimal node to some extent, it fails to suppress interference; the non-optimal nodes remain loosely scattered with noticeable outliers. The deeper model ($D=42, W=128$), however, exerts a strong compressive force on the background. As shown in Figure 9(c), it collapses non-optimal candidates into a dense, compact cluster, creating a clear margin that isolates the optimal node. This indicates that scaling depth specifically enhances separability, enabling the model to filter out background noise with high precision. 

\subsection{Feature Map Analysis}

\begin{figure}[ht]
\centering
\subfloat[Depth=6,Width=128]{\includegraphics[width = 0.33\linewidth]
{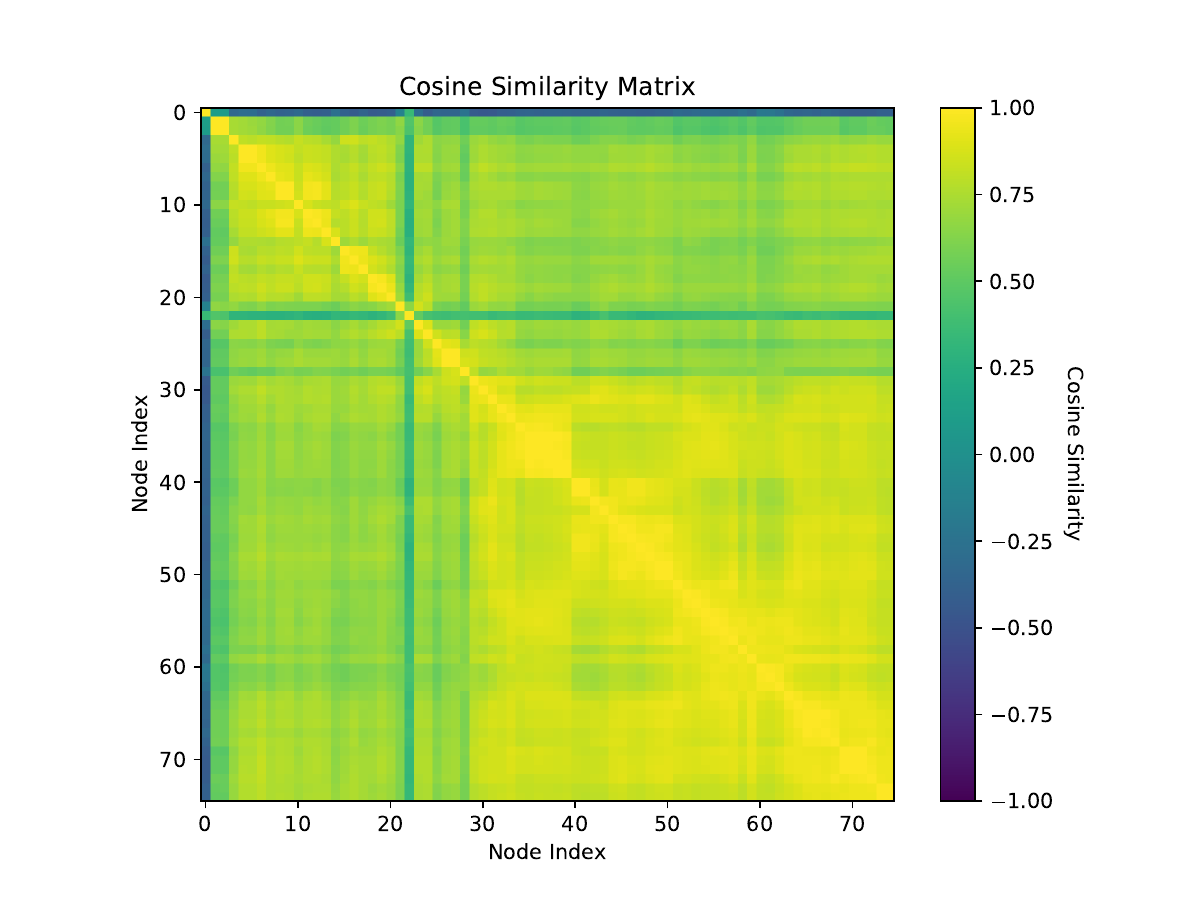}} \hspace{0mm}
\subfloat[Depth=6,Width=512] {\includegraphics[width = 0.33\linewidth]{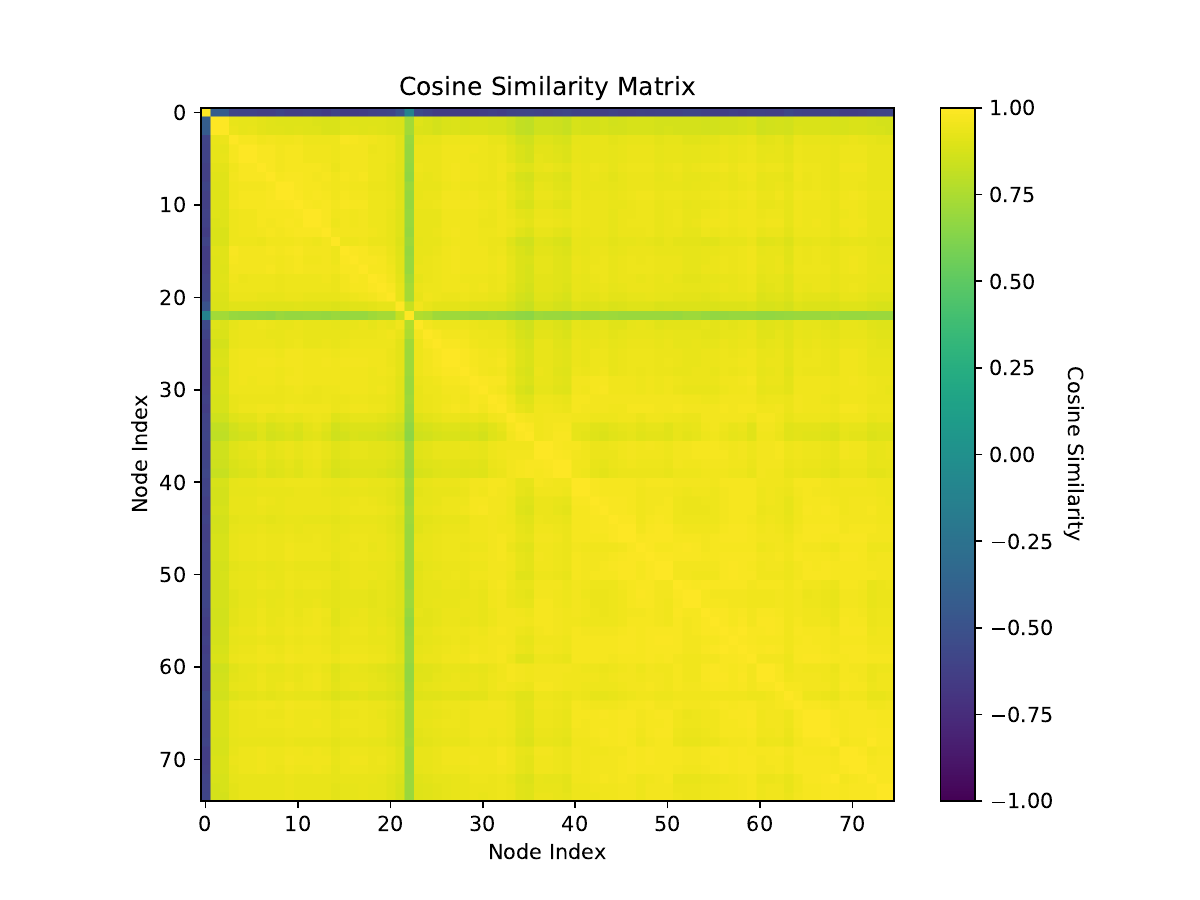}}\hspace{0mm}
\subfloat[Depth=42,Width=128]{\includegraphics[width = 0.33\linewidth]{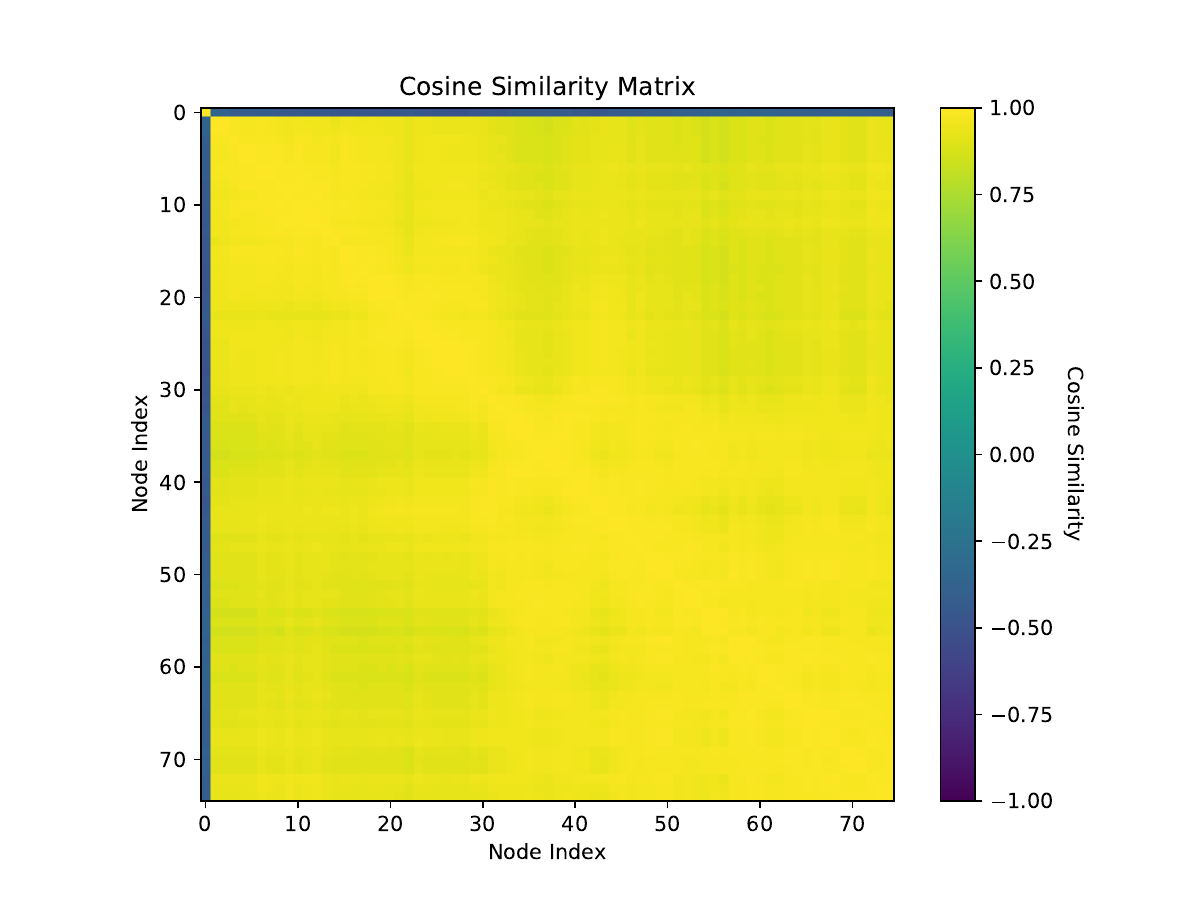}}
\caption{\textbf{Cosine similarity for all available node feature pairs. } Node indices are sorted according to the optimal solution. \textbf{(a) :} has distinct green blocks, indicating the presence of multiple clusters. \textbf{(b) :} The green area has noticeably become lighter, but has several distinct green lines, indicating the presence of a few outliers.  \textbf{(c) :} The single dark line is clearly distinguishable from the yellow block.} 
\label{fig: feature map}
\end{figure}

We observe a phenomenon consistent with the PAC results in our feature map analysis. The feature map is derived from the pairwise cosine similarity of all nodes, which are shown in Figure~\ref{fig: feature map}. For the baseline model ($D=6, W=128$), we observe numerous distinct green stripes, indicating the presence of multiple outliers within the latent space. In contrast, the deep model ($D=42, W=128$, 9.02M) exhibits a striking pattern: a dark line appears exclusively at index 0, while the remaining nodes form a homogeneous yellow block. This suggests that the model successfully isolates the optimal node while uniformly compressing the representations of non-optimal nodes. Although the wide model ($D=6, W=256$, 21.00M) shows a clearer clustering effect than the baseline, residual green stripes persist, implying insufficient spatial compression of interfering nodes. Consequently, we conclude that the scaling depth is superior for enhancing the model's discriminative capability.

\subsection{Study on Long-sightedness}

To further understand the mechanism behind the advantages of extended depth, we analyze the changes in the Long-sightedness capability as the model becomes wider or deeper: the ability to correctly identify the next optimal node even if it is relatively far away in space.

We evaluate this capability using 1,000 TSP100 instances, calculating the probability that the model correctly selects the optimal node when it corresponds to the $k$-th Euclidean neighbor of the current node. As Figure 8 illustrates, while performance naturally declines for all configurations as the target node becomes more distant, the rate of decay varies significantly. The baseline ($D=6, W=128$) and wider ($D=6, W=512$) models suffer substantial degradation, with accuracy dropping to $\sim$83\% and $\sim$87\% respectively for targets beyond the eighth neighbor ($k > 8$). In contrast, the deeper model ($D=42, W=128$) exhibits exceptional robustness, maintaining a success rate of over 93\% in these challenging scenarios. These results indicate that scaling depth minimizes local Euclidean biases; this enhanced long-sightedness ensures the precise selection of optimal nodes, directly translating into the generation of superior solutions.

\begin{figure}[ht]
    \centering
\centerline{\includegraphics[width=0.7\linewidth]{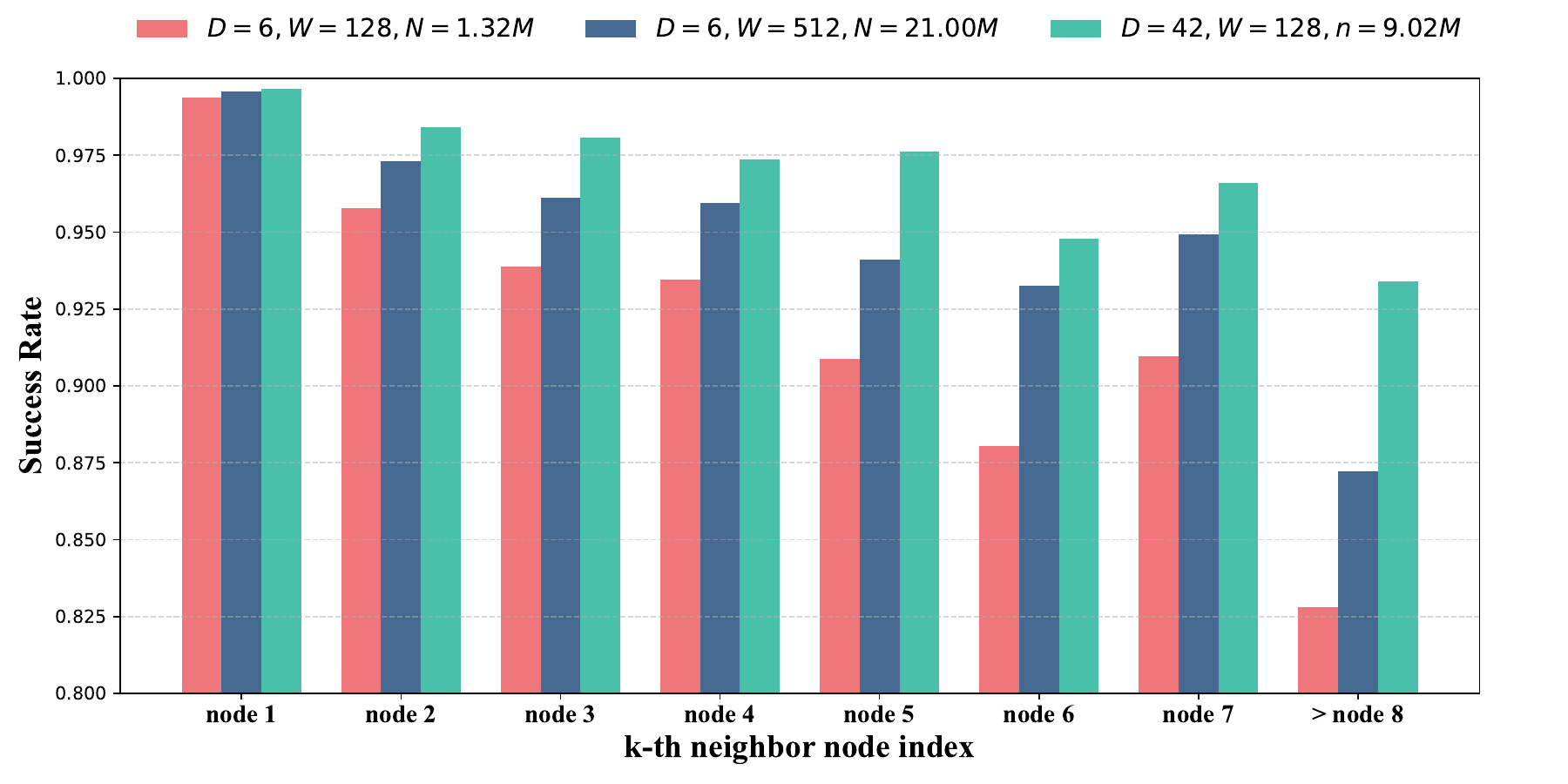}}
        \caption{The success rate when the optimal next node is located at different ranks of nearest neighbors.}
        \label{fig: long range robustness}
\end{figure}

\newpage

\section{Versatility of Design Principle}
\label{Versatility of Design Principle 1}

We empirically verify the versatility of Design Principle 1 by optimizing the representative heavy-decoder baseline LEHD model~\citep{luo2023neural}. We restructure the model into a deeper, narrower variant by reducing the embedding size from 128 to 64 for both the encoder and decoder. While the encoder depth remains unchanged, we significantly increase the decoder depth from 6 to 24 layers. As detailed in Table~\ref{tb: applied LEHD}, this architectural adjustment yields substantial improvements. Despite possessing $12.6\%$ fewer parameters (1.25M vs. 1.43M), the optimized model not only enhances in-domain performance (reducing the TSP100 gap by $27.4\%$) but also demonstrates significantly stronger out-of-domain generalization across TSP200--1000.
Most notably, on the large scale TSP1000, the optimality gap is reduced from $3.17\%$ to $1.84\%$, corresponding to a substantial $41.8\%$ relative improvement. This empirical evidence confirms the effectiveness of Design Principle 1 mentioned in Section~\ref{Parameter Placement Policy}.

\begin{table}[ht]
  \caption{Performance comparison of LEHD before and after modification on TSP100, TSP200, TSP500, and TSP1000.}
  \label{tb: applied LEHD}
  \begin{center}
    \begin{small}
      \resizebox{0.6\columnwidth}{!}{
        \begin{tabular}{ccccccc}
          \toprule
            Model (Parameters)   & TSP100  & TSP200 & TSP500 & TSP1000 \\
          \midrule
            LEHD $(D=6, W=128)$ (1.43M)  & 0.577\% & 0.859\%  & 1.560\% & 3.168\% \\
            LEHD $(D=24, W=64)$ (1.25M)  & \textbf{0.419\%}  & \textbf{0.597\%} & \textbf{1.025\%} & \textbf{1.844\%} \\
          \bottomrule
        \end{tabular}
        }
    \end{small}
  \end{center}
\end{table}

\end{CJK*}

\end{document}